\documentclass{article}

\PassOptionsToPackage{numbers,compress}{natbib}
\usepackage[final, main]{neurips_2026}

\usepackage[utf8]{inputenc}
\usepackage[T1]{fontenc}
\usepackage{hyperref}
\usepackage{url}
\usepackage{booktabs}
\usepackage{amsfonts}
\usepackage{nicefrac}
\usepackage{microtype}
\usepackage{graphicx}
\usepackage{dcolumn}
\usepackage{bm}
\usepackage{mathtools}
\usepackage{algorithm}
\usepackage{algpseudocode}
\usepackage{enumitem}
\usepackage{amssymb}
\usepackage{pifont}
\usepackage{multirow}
\usepackage{amsthm}
\usepackage{cancel}
\usepackage{chngcntr}
\usepackage{wrapfig}
\usepackage{caption} 
\usepackage{makecell}
\usepackage{siunitx}    
\usepackage{array}    
\usepackage{lipsum}
\usepackage{subcaption}

\newcommand{\coloneq}{\coloneqq}

\title{DiffeoMorph: Learning Distributed 3D Morphogenesis with Rotation-Invariant Spectral Shape Matching}

\author{%
  Seong Ho Pahng\textnormal{\textsuperscript{1, 2}}\thanks{Current address: Department of Physics, Brandeis University \ \ Email: \texttt{spahng@brandeis.edu}}
  \textnormal{\quad}
  Guoye Guan\textnormal{\textsuperscript{1, 2}} 
  \textnormal{\quad}
  Benjamin Fefferman\textnormal{\textsuperscript{2, 3, 4}} 
  \textnormal{\quad}
  Sahand Hormoz\textnormal{\textsuperscript{1, 2, 4}}\thanks{Corresponding author: \texttt{sahand\_hormoz@hms.harvard.edu}}
  \\
  \textsuperscript{1}Department of Systems Biology, Harvard Medical School
  \\
  \textsuperscript{2}Department of Data Science, Dana-Farber Cancer Institute 
  \\
  \textsuperscript{3}Department of Biomedical Informatics, Harvard Medical School
  \\
  \textsuperscript{4}Broad Institute of MIT and Harvard 
  \\
  \texttt{\{spahng@ds.dfci, guoye@ds.dfci, bfefferman@g, sahand\_hormoz@hms\}.harvard.edu}
}

\begin{document}

\maketitle

\begin{abstract}
Biological systems can form complex three-dimensional structures through the collective behavior of agents that share a common update rule and operate without central control. How such distributed control gives rise to precise global patterns remains a central question not only in developmental biology but also in distributed robotics, programmable matter, and multi-agent learning. Here, we introduce DiffeoMorph, an end-to-end differentiable framework for learning a morphogenesis protocol that guides a population of agents to morph into a target 3D shape. Each agent updates its position and internal state using an SE(3)-equivariant graph neural network, based on its own internal state and signals received from other agents. To train this system, we introduce a new shape-matching loss based on 3D Zernike polynomials, which compares the predicted and target shapes as continuous spatial distributions, not as discrete point clouds, and is invariant to agent ordering, number of agents, and global orientation. To achieve rotation invariance while preserving reflection sensitivity, we include an alignment step that optimally rotates the predicted Zernike spectrum to match the target before computing the loss. We perform benchmarking to establish the advantages of our shape-matching loss over other standard distance metrics for shape comparison tasks. We then demonstrate that DiffeoMorph can form a range of complex shapes from minimally patterned initial conditions. DiffeoMorph provides a general framework for learning distributed control strategies for morphogenesis, swarm robotics, and programmable self-assembly. Code is available on  \href{https://github.com/hormoz-lab/diffeomorph}{GitHub}.

\end{abstract}

\section{Introduction}

In biology, identical cells---each executing the same internal regulatory program---can coordinate to form precise and complex three-dimensional structures \citep{gilbert2020developmental}. These cells interact through a combination of contact-mediated adhesion \citep{takeichi2014dynamic} and long-range signaling via diffusible morphogens \citep{gurdon2001morphogen}, yet no cell has global knowledge of the entire system. This raises a fundamental question: how can a single model, shared across all agents, give rise to such intricate collective behavior? This question lies at the heart of developmental biology, and answering it could transform our ability to engineer artificial systems, from self-assembly \citep{whitesides2002self} and swarm robotics \citep{rubenstein2014programmable} to organoids \citep{fatehullah2016organoids}.

Machine learning offers a compelling path to learning such rules automatically, rather than designing them by hand. Two main paradigms have been explored. The first is reinforcement learning (RL) \citep{sutton1998reinforcement}, which optimizes decentralized agent policies using reward signals tied to shape formation \citep{lin2020modeling,viquerat2021direct,deshpande2025engineering}. However, these methods tend to be inefficient due to the lack of end-to-end differentiability. The second paradigm is end-to-end differentiable learning, successfully adopted in Neural Cellular Automata (NCA) \citep{mordvintsev2020growing,grattarola2021learning} to learn morphogenetic rules via gradient-based optimization.

However, all of these approaches ultimately rely on comparing a simulated shape to a target shape, and existing formulations, which compare shapes using a Cartesian coordinate matrix \( \mathbf{X} {\in} \mathbb{R}^{N {\times} 3}\), suffer from one or more of the following constraints relative to the target: equal number of agents, pointwise correspondence, and identical global orientation. These assumptions are often not satisfied in both natural and engineered multicellular systems, and thus the shape-matching loss should be free of these requirements if it is intended for training autonomous morphogenesis rules in realistic settings.

To overcome these limitations, we introduce a spectral shape-matching loss based on the 3D Zernike polynomials \citep{niu2022zernike, novotni2004shape}, which consist of spherical harmonics weighted by radial eigenfunctions. We represent both the generated and target shapes as distributions in space and compare their spectral signatures---the expansion coefficients with respect to Zernike polynomials. This yields a loss function that is \textit{invariant to agent labeling}, \textit{robust to differences in the number of agents}, and \textit{agnostic to absolute position}. To further ensure \textit{rotation invariance}, we include a spectral alignment step that learns the 3D rotation—parameterized by a unit quaternion—that optimally rotates the spectrum of the generated shape to match that of the target before computing the loss. The loss retains the full spectrum during the alignment step and thus preserves \textit{sensitivity to reflection}, an essential property for accurately capturing complex morphologies with chirality.

We combine this novel shape-matching loss with an SE(3)-equivariant force model to introduce \textbf{DiffeoMorph}, an end-to-end differentiable framework for learning morphogenesis control that drives agents to collectively form target 3D shapes. During simulation, the force model updates each agent's position and internal states based on its current configurations and the signals received from other agents. Because the loss internally optimizes the alignment between the predicted and target Zernike spectra, each gradient step over model parameters requires solving the alignment optimization problem. This dependency yields a bilevel optimization structure: the outer loop updates the agent model, while the inner loop solves for optimal spectral alignment. We show that the alignment optimization needs to be computed only in the forward pass, and that the gradient through this inner optimization can be omitted during backpropagation, because the aligned loss is locally stationary with respect to rotations; in practice, we use a lightweight Riemannian implicit-differentiation correction to account for the finite accuracy of the inner optimization. We demonstrate that, starting from an uninformative spherical configuration with only minimal spatial cues encoded in the agents' internal states, DiffeoMorph can robustly generate a range of complex 3D shapes.
The main contributions of this paper are the following:
\begin{enumerate}[label=\textbf{\arabic*.}, leftmargin=*]
    \item \textbf{Novel loss function.} We introduce a spectral shape-matching objective with an intermediate alignment step, achieving invariance to permutation, population size, and global orientation while preserving chirality. We show that the gradient through the alignment optimization can be omitted during backpropagation for an exact solve, and use a lightweight Riemannian implicit-differentiation correction for finite-accuracy solves.
    
    \item \textbf{Morphogenesis learning framework.} We propose DiffeoMorph, a differentiable agent-based framework that trains a shared SE(3)-equivariant dynamics model using the proposed loss to learn morphogenesis rules that drive the self-organization of agents into complex 3D shapes.
    
    \item \textbf{Empirical analysis.} We demonstrate that our loss uniquely satisfies all desired properties while achieving competitive runtime against standard benchmarks for shape comparison. We systematically conduct shape learning experiments and analyze the learned internal states to identify fundamental requirements and candidate organizing principles underlying morphogenesis.

\end{enumerate}

\section{Methodological Framework}
\label{section:framework}
\begin{figure}[t!]
    \centering
    \includegraphics[width=1\linewidth]{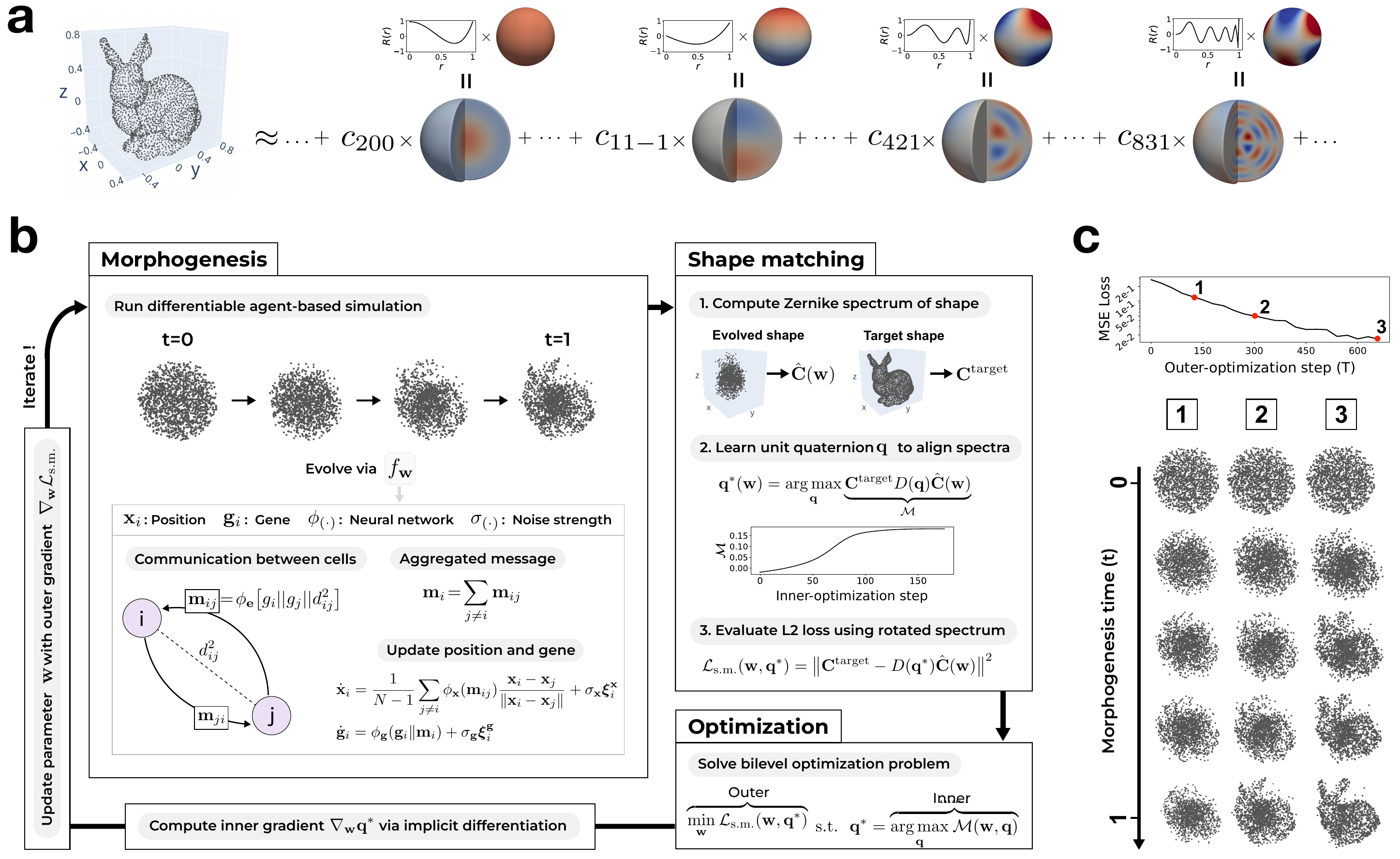}
    \caption{\textbf{Schematic of DiffeoMorph.} (a) A 3D shape represented as a point cloud can be expressed as a linear combination of the 3D Zernike polynomials, composed of radial and angular parts. The expansion coefficients \( \mathbf{C} {\coloneq}  \{c_{n\ell m} \} \) are the Zernike moments. (b) The morphogenesis force model evolves the positions and gene expressions of agents based on their gene expressions and distances to neighbors, without access to absolute position information. The final evolved shape is compared to a target shape using their spectra given by 3D Zernike moments. A bilevel optimization procedure aligns the spectra by learning the unit quaternion (inner optimization) and updates model parameters to minimize the shape‐matching loss (outer optimization). (c) As the outer optimization proceeds, the simulation produces the desired target shape.}
    \label{main:fig1}
\end{figure}

Figure~\ref{main:fig1} gives an overview of the DiffeoMorph framework. In the following sections, we describe its components and how they fit together to support morphogenesis learning.

\subsection{Shape-Matching Objective based on 3D Zernike Polynomials}

\subsubsection{Spectral representation of 3D shapes}
\label{section:loss:spectral_representation}
Consider a point cloud representing the positions of \( N \) agents evolved by an agent-based model, denoted by \( \mathbf{X}^\text{evol} {\in} \mathbb{R}^{N {\times} 3} \) and expressed in a chosen Cartesian laboratory frame. Our goal is to compare it with a target shape \( \mathbf{X}^\text{target} {\in} \mathbb{R}^{M {\times} 3}\) where \( M\) may differ from \( N \) and, even when \(M{=}N\), the point-to-point correspondence between two shapes may not be known. To perform shape comparison in a permutation-invariant and agent-number-agnostic manner, we seek to treat these matrices as discrete approximations of functions over space and expand them in suitable basis functions.

To this end, we first constrain \(\mathbf{X}^\text{evol} \) to lie within the unit ball \( B_1(0) \) as \( \mathbf{X}^\text{evol}_B {=} ( \mathbf{X}^\text{evol} {-} \mathbf{X}^\text{evol}_\text{c.o.m} ) / r_\text{max} \) where \( \mathbf{X}^\text{evol}_\text{c.o.m}\) denotes the center-of-mass coordinates broadcasted over all rows of \(\mathbf{X}^\text{evol}\) and \( r_{\max} \) is the largest radius of the centered target point cloud. For clarity, we omit the superscript in what follows. We define the following normalized empirical measure supported in the unit ball \( B_1(0) \), \(\mu(\mathbf{x}) {=} \frac{1}{N} \sum_{i=1}^N \omega_i  \delta(\mathbf{x} {-} \mathbf{x}_{B,i})
\), where we allow each point to be associated with a distinct learnable weight \( \omega_i \). We expand \(\mu \) in real 3D Zernike polynomials \( Z_{n \ell m} \)---which form an orthonormal basis of \( L^2 \left( B_1 (0) \right) \)---with truncation parameters \(n_{\max} \) and \( \ell_{\max} \) as,
\begin{align}
  \mu(\mathbf{x}) & \approx 
    \sum_{n=0}^{n_{\text{max}}} 
    \sum_{\ell=0}^{\ell_{\text{max}}} 
    \sum_{m=-\ell}^{\ell}
      c_{n\ell m}
      \underbrace{
        \overbrace{
          R_{n\ell}(r)
        }^{\text{radial}}
        \, 
        \overbrace{
          Y_{\ell m}^{\text{real}}(\theta, \phi)
        }^{\text{angular}}
      }_{Z_{n\ell m}(\mathbf{x})}
      \label{eq1}
  \\
  \text{where} 
  \quad
  c_{n \ell m} &=
    \langle  \mu, Z_{n\ell m} \rangle =
    \frac{1}{N}\sum_{i=1}^N
      \omega_i 
      R_{n\ell}(r_i) 
      Y_{\ell m}(\theta_i, \phi_i).
      \label{eq2}
\end{align}
In this functional representation, the simulated 3D shape is registered via the coefficients of the 3D Zernike polynomials, or Zernike moments, \( \mathbf{C} {\coloneqq} \{c_{n \ell m} \} \), obtained from the projection \eqref{eq2}. These can be viewed as a spectrum that contains both radial and angular variation of the shape. Figure~\hyperref[main:fig1]{1a} visualizes a few terms of the expansion of the bunny, in the order of increasing complexity of the 3D Zernike polynomials, illustrating the transformation from the Euclidean to the spectral representation. Further mathematical properties of the 3D Zernike polynomials are discussed in Appendix~\ref{supp_sec:zernike}. 

The summation over all agents \(i\) in the projection, the normalization in the definition of the empirical measure, and the subtraction of the center-of-mass position ensure invariance to agent index permutation, the total number of agents, and global translation, respectively. However, Zernike moments \( \mathbf{C} \) remain sensitive to orientation because the azimuthal index \( m \) of the spherical harmonics depends on the orientation of the shape with respect to the coordinate axes of the laboratory frame. Consequently, the Zernike moments of a rotated shape differ from those of the original.

\subsubsection{Spectral alignment for rotation invariance and chirality sensitivity}
\label{section:loss:spectral_alignment}

Instead of collapsing the azimuthal indices \( m\), either by computing the norm \citep{novotni20033d} or higher-order spectra \citep{hu2001angular}, which would compromise the discriminative power of spectral representation or require additional computations, we seek an alternative approach where we align the spectra of two shapes before direct comparison in order to achieve rotation invariance. Given two sets of Zernike moments \( \mathbf{C} \) and \( \tilde{\mathbf{C}} \) computed from two 3D shapes with different orientations but otherwise the same overall geometry, Wigner--D matrices \( D^\ell {\in} \mathbb{R}^{(2\ell+1) {\times}  (2\ell +1)} \) at each angular degree \( \ell \) can rotate one into the other according to \( \tilde{c}_{n \ell m} {=} \sum_{m'} D^\ell_{m m'} c_{n \ell m'}\), which holds for all radial orders \( n \) as \( R_{n \ell} \) does not depend on rotation. We choose unit quaternions rather than Euler angles to parametrize rotations, because they provide a smooth, singularity-free representation of \(\mathrm{SO}(3)\) and avoid the gimbal-lock issues inherent to Euler angles. Additional details of the procedure, comparison with the Euler-angle formulation, and relevant mathematical background are presented in Appendix~\ref{supp_sec:representation}; the quaternion-polynomial expression for the complex Wigner--D matrix and the corresponding real-valued Wigner--D matrix used throughout this work are given in Equations~\eqref{supp:eq5} and~\eqref{eq:real_valued_wigner}, respectively.

Let \( \varphi_\mathbf{w} \) denotes a chosen morphogenesis model parameterized by weight \( \mathbf{w}\), which evolves the initial configuration \(\mathbf{X^0}\) into the evolved configuration \(\mathbf{X^\text{evol}}\), and \(\Phi\) the function that constrains a point cloud to the unit ball and projects it onto the 3D Zernike polynomials---such that \( \mathbf{C}^\text{evol} {=} \Phi \left( \mathbf{X}^\text{evol}\right) \) and \( \mathbf{X}^\text{evol} {=} \varphi_\mathbf{w} \left(\mathbf{X^0} \right) \). We can formulate the desired shape-matching objective for learning parameters \(\mathbf{w}\) as the outer loss of the following bilevel optimization problem,
\begin{equation}
  \mathcal{L}_{\text{s.m.}} \! \left(\mathbf{w};  \mathbf{X}^0 \right) = \frac{1}{N_\text{spec}}
    \sum_{n=0}^{n_\text{max}} 
    \sum_{\ell=0}^{\ell_\text{max}}
    \sum_{m=-\ell}^\ell
      \big\| 
        c^\text{target}_{n \ell m} - 
        \sum_{m'=-\ell}^\ell \! D^\ell_{mm'}\big(\mathbf{q}^\star(\mathbf{w})\big) \, 
        c^\text{evol}_{n \ell m'}
      \big\|^2 
    + \big\| \mathbf{x}^\text{evol}_\text{c.o.m}\big\|^2
\label{eq5}
\end{equation}
where \(N_\text{spec}\) is the total number of \((n, \ell, m)\) spectral index configurations appearing in the expansion and \(\mathbf{x}^\text{evol}_\text{c.o.m} \) is the center-of-mass coordinates, and the optimal unit quaternion \( \mathbf{q}^\star(\mathbf{w}) \) for a given \( \mathbf{w} \) is found in the inner optimization,
\begin{equation}
\mathbf{q}^\star(\mathbf{w}) =
\operatorname*{\arg\max}_{\mathbf{q\in S^3}}
\underbrace{
\frac{1}{N_\text{spec}}
\sum_{n=0}^{n_\text{max}} 
\sum_{\ell=0}^{\ell_\text{max}}
\sum_{m,m'=-\ell}^\ell
c^\text{target}_{n \ell m} \,
D^\ell_{mm'}(\mathbf{q}) \,
c^\text{evol}_{n \ell m'}
}_{\mathcal{M}(\mathbf{q}\,;\mathbf{w})}
\label{eq6}
\end{equation}
which we refer to as the spectral overlap as it maximizes the alignment between the rotated and the target spectra, and solve using gradient updates as there is no closed-form solution. By retaining all azimuthal components \(m\), the shape-matching objective remains rotation invariant while distinguishing chirality. The center-of-mass correction term prevents the model from exploiting coordinated drift, ensuring that shape formation arises from genuine deformation. Details of the Riemannian gradient optimization of a unit quaternion are provided in Appendix~\ref{supp_sec:riemannian_gradient}. Figure~\ref{supp:quaternion_alignment} shows that quaternion-based optimization
can align shapes when Euler-angle optimization fails due to gimbal lock.

\subsubsection{Differentiation of the shape-matching objective}
\label{section:loss:implicit_differentiation}

Differentiating the aligned outer loss \(\mathcal{L}_\text{s.m.}(\mathbf{w},\mathbf{q}^\star(\mathbf{w}))\) formally gives \(\nabla_\mathbf{w}\mathcal{L}_\text{s.m.}{+}\nabla_\mathbf{q}\mathcal{L}_\text{s.m.}\nabla_\mathbf{w}\mathbf{q}^\star\). In a generic nested optimization problem, the second term matters: as \(\mathbf{w}\) changes, the optimal inner solution \(\mathbf{q}^\star\) changes as well, requiring either unrolled backpropagation or implicit differentiation. Our aligned spectral loss has a simpler structure. The inner alignment objective is exactly the \(\mathbf{q}\)-dependent part of the outer spectral MSE. The Wigner-D matrices are orthogonal, \({D^\ell(\mathbf{q})}^{\!\top}\!D^\ell(\mathbf{q}){=}I\), so rotating the predicted spectrum changes its orientation but not its squared norm. When the MSE is expanded, both the target norm and the predicted-spectrum norm are independent of \(\mathbf{q}\); the only \(\mathbf{q}\)-dependent term is the cross term between the target spectrum and the rotated predicted spectrum. This cross term is precisely the spectral overlap \(\mathcal{M}\) maximized by the alignment step.

Thus, at an exact alignment optimum, infinitesimal tangent perturbations of the unit quaternion do not change the aligned loss to first order. Equivalently, the Riemannian gradient of the loss with respect to the constrained variable \(\mathbf{q}{\in} S^3\) vanishes: \(\nabla^{S^3}_\mathbf{q}\mathcal{L}_{\mathrm{s.m.}}(\mathbf{w},\mathbf{q}^{\star}){=}0\). Although \(\mathbf{q}^\star\) depends on \(\mathbf{w}\), its movement is along directions in which the loss is locally stationary. Therefore the tangent-space term \(\nabla^{S^3}_\mathbf{q}\mathcal{L}_\text{s.m.}\nabla_\mathbf{w}\mathbf{q}^\star\) vanishes at an exact local optimum, and the outer gradient reduces to \(\nabla_\mathbf{w}\mathcal{L}_{\mathrm{s.m.}}(\mathbf{w},\mathbf{q}^{\star}(\mathbf{w})){=}\nabla_\mathbf{w}\mathcal{L}_{\mathrm{s.m.}}(\mathbf{w},\mathbf{q}^{\star})\). This is the standard envelope-theorem property of a locally optimized inner variable. Thus, for a converged alignment solve, the correct first-order gradient does not require differentiating through the inner optimization, and \(\mathbf q^\star\) can be detached from the computational graph during the backward pass with respect to \(\mathbf w\).

In practice, the inner alignment is solved numerically with a finite stopping threshold and maximum iteration cutoff, producing an approximate optimizer \(\hat{\mathbf{q}}\). Then \(\nabla^{S^3}_\mathbf{q}\mathcal{L}_{\mathrm{s.m.}}(\mathbf{w},\hat{\mathbf{q}})\) is small but not exactly zero, so the skipped term is small but not identically absent. We therefore implement Riemannian implicit differentiation as an optional correction for this residual dependence of the approximate alignment on \(\mathbf{w}\). This correction avoids unrolling the inner optimization and is inexpensive because the alignment variable is only a unit quaternion. In Figure~\ref{supp:runtime}, we show that implicit differentiation has negligible runtime overhead relative to detaching the alignment, while providing a substantial speedup over unrolled backpropagation. Appendix~\ref{supp_sec:implicit_differentiation} derives this Riemannian implicit-differentiation term and shows how it provides a first-order correction when the inner alignment is only approximately solved.

\subsection{Morphogenesis Model}
\label{section:model}

\paragraph{State of an Agent.} Drawing inspiration from biological cells, we assign to each agent \( i \) a \( d_\mathbf{g} \)-dimensional gene expression vector \( \mathbf{g}_i {\in} \mathbb{R}^{d_\mathbf{g}} \). Across the population, they are aggregated into a matrix \( \mathbf{G} \) and evolve together with the position matrix \( \mathbf{X} \) during the morphogenesis simulation. By construction, the gene expression vector \( \mathbf{g}_i \) encodes rotation-invariant features. At each time step \(t\), the system is described by the positions of the agents \(\mathbf{X}^t\) and their gene expression \(\mathbf{G}^t\). For clarity, we omit the time index \(t\) in what follows.

\paragraph{SE(3)-equivariant force model.} 
To evolve the states of agents in a manner consistent with rigid body transformations, we employ an SE(3)-equivariant EGNN \citep{satorras2021n} to compute the forces for time integration. \(\phi_{(\cdot)}\) in the description below denotes a multi-layer perceptron (MLP) and \( \mathbf{w} \) collectively denotes their learnable parameters. For each pair of agents \(i\) and \(j\), we first construct the edge feature \(\mathbf{e}_{ij} {=} [\mathbf{g}_i \Vert \mathbf{g}_j \Vert d_{ij}^2]\), where \(d_{ij}^2\) is the squared Euclidean distance. This is transformed into a message \(\mathbf{m}_{ij} {=} \phi_e(\mathbf{e}_{ij})\), capturing the communication between agents. Based on these messages, the dynamics governing the evolution of positions and gene expressions are given by
\begin{equation}
\begin{alignedat}{2}
\dot{\mathbf{x}_i} = \mathbf{f}_i^\mathbf{x} &= \frac{1}{N-1}\sum_{j \neq i}\phi_{\mathbf{x}}(\mathbf{m}_{ij}) \frac{\mathbf{x}_i - \mathbf{x}_j}{\Vert\mathbf{x}_i - \mathbf{x}_j\Vert}\,  + \sigma_\mathbf{x} \boldsymbol{\xi}_i^\mathbf{x} \\
\dot{\mathbf{g}_i} = \mathbf{f}_i^{\mathbf{g}} &= \phi_\mathbf{g} \left(\mathbf{g}_i \Vert \mathbf{m}_i \right) + \sigma_\mathbf{g} \boldsymbol{\xi}_i^\mathbf{g} \quad \text{where} \quad \mathbf{m}_i = \sum_{j \neq i}\mathbf{m}_{ij} \\
\end{alignedat}
\label{eq11}
\end{equation}
where the dot notation denotes the time derivative, and \( \sigma_{\mathbf{x}/\mathbf{g}}\) control the noise strengths and \( \boldsymbol{\xi}_{(\cdot)}^{\mathbf{x}/\mathbf{g}}\) are corresponding Gaussian white noises for positions and gene expressions, respectively. Pseudocode for the force computation steps is provided in Appendix~\ref{supp_sec:pseudocode}.

\section{Related Work}
The alignment step in our loss can be viewed as shape registration—finding a transformation that aligns one shape to another. When pointwise correspondences are known, this reduces to Orthogonal Procrustes \citep{schonemann1966generalized,gower2004procrustes}. Without correspondences, Iterative Closest Point \citep{besl1992method} or Sinkhorn-based matchings \citep{cuturi2013sinkhorn,wang2019deep} can be used to estimate them. Many shape-matching objectives have been proposed in computer vision. For point clouds, Chamfer Distance and Earth Mover's Distance are widely used to quantify geometric discrepancy \citep{fan2017point,rubner2000earth}. For implicit representations such as Occupancy Networks \citep{mescheder2019occupancy} or Signed Distance Functions \citep{osher1988fronts,park2019deepsdf}, Binary Cross Entropy or functional norms are commonly employed. These objectives, however, depend on ambient-space sampling and are sensitive to rigid transformations and point density. Relational Metrics such as the Gromov--Wasserstein distance \citep{memoli2011gromov,peyre2019computational} avoid the need for pointwise correspondences and are rotation invariant by comparing the pairwise distance matrices of the two shapes. However, they are computationally expensive and cannot distinguish chirality. Spectral approaches based on Zernike polynomials—widely used in optics, imaging, and astronomy \citep{von1934beugungstheorie,teague1980image,boland2001neural,alizadeh2016measuring,noll1976zernike,capalbo2021three}—offer a viable alternative. Prior works achieve rotation invariance by summing over the azimuthal index \(m\) (power spectrum) \citep{kazhdan2003rotation,novotni20033d,novotni2004shape} or by constructing higher-order invariants through angular couplings using Wigner 3j or 6j symbols (bispectrum and trispectrum) \citep{scoccimarro2000bispectrum,kakarala2012bispectrum,collis1998higher}. However, these approaches eliminate the directional information carried by the azimuthal index \(m\) and thus lose sensitivity to chirality. As described below, we benchmark our aligned full-spectrum Zernike objective against these point-cloud, relational, and spectral alternatives, and show that it uniquely combines permutation and point-count robustness, rotation invariance, chirality sensitivity, and favorable computational scaling.

Deep learning approaches to shape generation include RL \citep{deshpande2025engineering}, ranging from centralized agents with full shape information access \citep{lin2020modeling,viquerat2021direct} to decentralized multi-agent systems that self-assemble into target morphologies \citep{pathak2019learning}. However, the lack of end-to-end differentiability leads to inefficient training, particularly when shape formation depends on internal agent states. In differentiable settings, Neural Cellular Automata (NCA) have demonstrated the ability to generate structured patterns in both 2D \citep{mordvintsev2020growing,palm2022variational} and 3D domains \citep{zhang2021learning}. Graph NCA (GNCA) \citep{grattarola2021learning} and Neural Particle Automata (NPA) \citep{kim2026neural} generalize this paradigm to irregular and continuous space using graph neural networks and Smoothed Particle Hydrodynamics (SPH) kernel operators, respectively. Despite these architectural advances, training setups in these studies rely on informative initial conditions, non-equivariant architectures, or losses requiring known pointwise correspondences or fixed global orientations. 

\section{Experiments}
\subsection{Testing Shape-Matching Objective}
\label{section:loss_testing}
\begin{figure}[t!]
    \centering
    \includegraphics[width=1\linewidth]{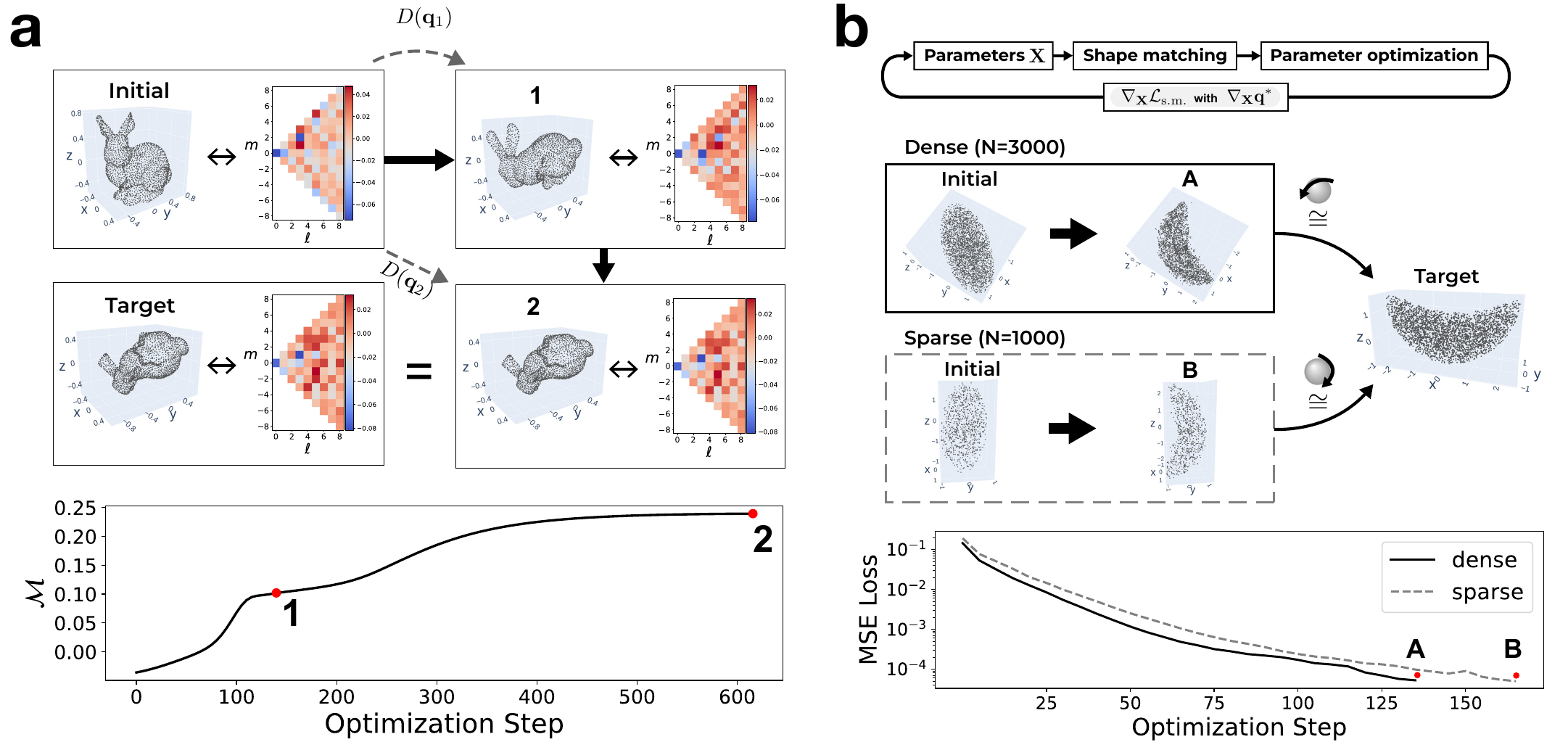}
    \caption{\textbf{Spectral shape-matching loss.} (a) The spectra of two shapes, given by the Zernike moments, can be aligned by solving for the unit quaternion that parametrizes the Wigner-D matrix maximizing the spectral overlap \(\mathcal{M}\). (b) To isolate the behavior of the shape-matching loss, we bypass the simulation step and directly optimize the input point cloud \( \mathbf{X} \). As the loss value is minimized, both dense and sparse ellipsoidal point clouds—each with a different initial orientation—successfully morph into the target shape up to rotation.}
    \label{main:fig2}
\end{figure}

Because the shape-matching loss of DiffeoMorph can be integrated with arbitrary force models or used in different learning paradigms such as RL, we begin by decoupling the learning problem from the simulation component to evaluate the behavior of our loss in a controlled setting. We designed two proof-of-concept experiments to test whether the rotation aligning two shapes in spectral space can be learned via gradient descent (inner optimization) and to verify whether shapes can be learned up to rotation due to the intermediate spectral alignment step (outer optimization). 

\paragraph{Spectral Alignment.} 
\label{section:result:spectral_alignment}
As a test shape, we used an unweighted point cloud of the Stanford bunny normalized to lie within the unit ball. The expansion coefficients, Zernike moments \(\mathbf{C}\), up to angular degree \(\ell{=}8\) are visualized as a matrix in Figure~\hyperref[main:fig2]{2a} at radial order \(n{=}1\). The goal is to find the optimal unit quaternion \( \mathbf{q} \) whose Wigner--D matrices \( D^\ell \) align the Zernike moments of the upright bunny shown in the ``Initial" box to those of the toppled bunny in the ``Target" box. As the spectral alignment progresses from point 1 to point 2 in Figure~\hyperref[main:fig2]{2a}, the predicted spectrum rotates to match the target. Applying the corresponding spatial rotation matrix \( R \) to the point cloud yields the toppled bunny, demonstrating that inner optimization can learn the optimal \( \mathbf{q} \) by maximizing the spectral overlap \( \mathcal{M} \).

\paragraph{Direct Shape Learning.}
\label{section:result:direct_shape_learning}
To assess whether our loss induces an informative gradient signal, we bypass integrating the morphogenesis model and we directly optimize the point cloud \(\mathbf{X}\) by minimizing the shape-matching loss \( \mathcal{L}_{\text{s.m.}} \) via its gradient \( \nabla_\mathbf{X} \mathcal{L}_{\text{s.m.}} \), as illustrated in the top row of Figure~\hyperref[main:fig2]{2b}. This experiment remains relevant to the full DiffeoMorph framework, as optimizing the agent model parameters ultimately involves backpropagating through \(\mathbf{X}\) via the chain rule: \( \nabla_\mathbf{w} \mathcal{L}_{\text{s.m.}} {=} \nabla_\mathbf{X} \mathcal{L}_{\text{s.m.}}  \nabla_\mathbf{w} \mathbf{X} \). The goal in this setup is to morph two initial ellipsoidal point clouds—differing in orientation and number of agents—into a target shape with different morphology and cardinality. As shown in Figure~\hyperref[main:fig2]{2b} (bottom), both point clouds at convergence—denoted by points “A” and “B”—successfully replicate the crescent shape despite retaining their initial orientations. This illustrates the rotation invariance of our objective: optimization converges once the correct morphology is 
recovered up to rotation. The inner alignment step---represented by a 3D rotation glyph---correctly matches spectra across orientations, yielding low MSE even when raw point clouds differ by rotation. These results in turn demonstrate that the gradients for the outer optimization are accurately computed via implicit differentiation. Figure~\ref{figure:weighted_shape} confirms that spectral alignment extends to
weighted shapes and that the weights \(\boldsymbol{\omega}\) can be learned
jointly with \(\mathbf X\).

\paragraph{Benchmarking shape-matching objective.}
\label{section:result:loss_benchmarking}
Since the shape-matching loss in DiffeoMorph is a key contribution, we benchmark our loss against standard distance metrics for shape comparisons: Chamfer distance \citep{fan2017point}, Earth Mover’s distance \citep{rubner2000earth}, Pairwise distance \citep{gala2024n}, Gromov–Wasserstein distance \citep{memoli2011gromov}, Power Spectrum \citep{kazhdan2003rotation, novotni20033d, novotni2004shape}, bispectrum \citep{scoccimarro2000bispectrum, kakarala2012bispectrum, collis1998higher}, and trispectrum \citep{collis1998higher}. As shown in Figures~\hyperref[figure:loss_benchmarking]{S4a,b,d}, our loss uniquely satisfies invariance to permutation, point count, and rotation while remaining sensitive to chirality, and achieves superior runtime scaling with respect to the number of points. We further perform direct shape learning experiments with chirality-insensitive metrics in Figure~\hyperref[figure:loss_benchmarking]{S4c}, confirming that they fail to recover the correct head orientation of the Stanford bunny.

\subsection{Learning to morph 3D shapes via agent-based simulation}

\begin{figure}[t!]
    \centering
    \includegraphics[width=1\linewidth]{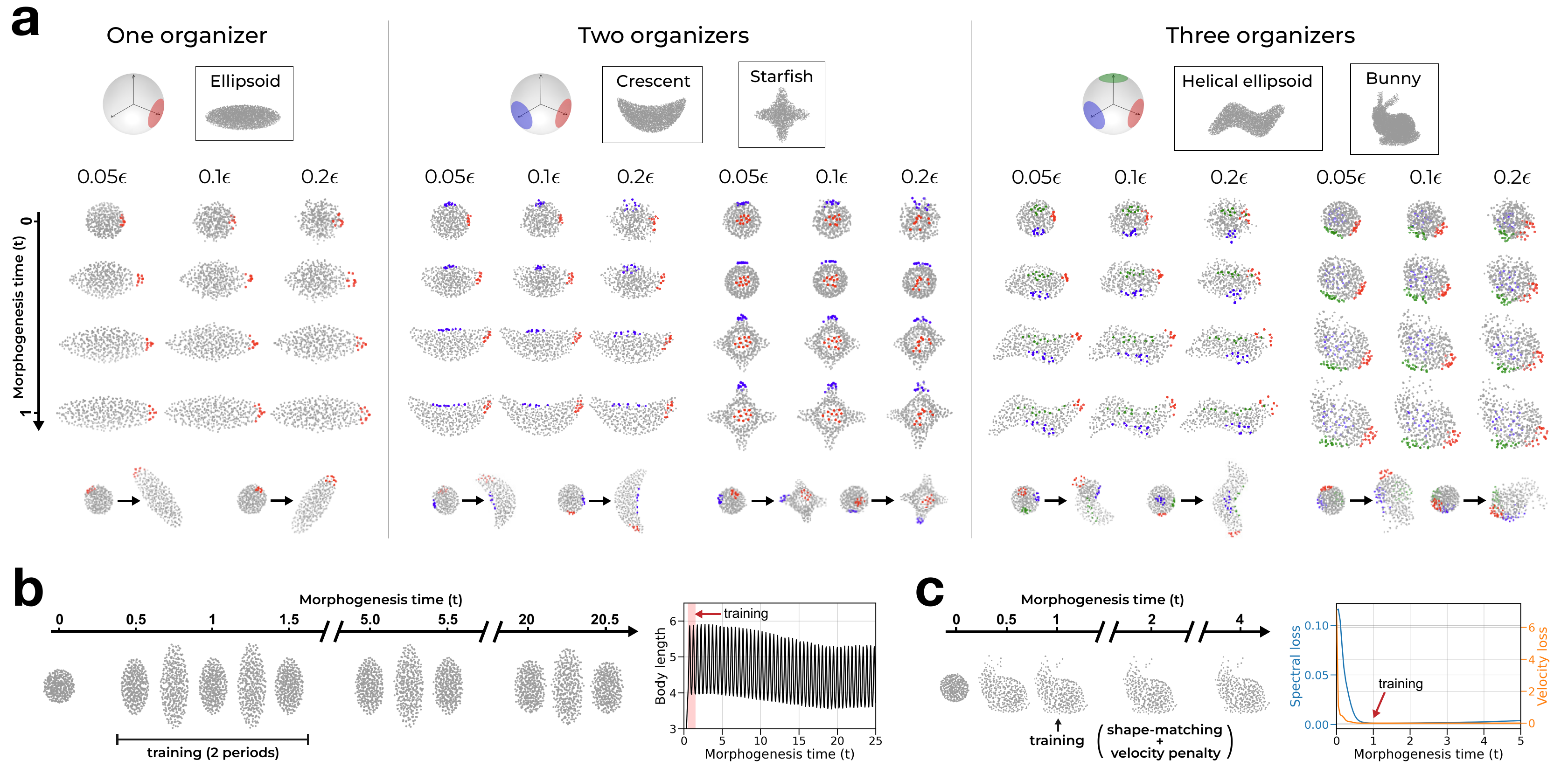}
    \caption{\textbf{Visualization of morphogenesis trajectories.} (a) Colored regions denote “organizer cells” that provide minimal spatial cues for morphogenesis. $\epsilon$ denotes zero-mean Gaussian noise. The trained model generalizes to higher noise at the training orientation and to organizer rotations. (b) Shape matching can occur at multiple times. By training over two periods with alternating long and short ellipsoid targets, the model learns to oscillate between these forms well beyond training horizon. (c) Including a velocity penalty at the final time encourages the model to form a stationary shape.} 
    \label{main:fig4}
\end{figure}

To demonstrate the full capabilities of DiffeoMorph, we evaluate its performance in morphing a population of agents into a curated set of 3D target shapes with progressively increasing geometric complexity: an ellipsoid, a crescent, a starfish, a helical ellipsoid, and the Stanford bunny. 

Before presenting results, we highlight a key design consideration. If each agent were initialized with a unique internal state---for example, by encoding its initial spatial coordinates \([x_i, y_i, z_i]\) or any other agent-specific identifier---and if the morphogenesis simulation contained no noise, then a simple MLP acting independently on each agent could trivially learn a pointwise mapping from the initial cloud to the target. In this regime, morphogenesis degenerates into trivial coordinate regression and requires no collective behavior. To make the task biologically and algorithmically meaningful, we deliberately withhold initial spatial information and provide only minimally informative internal states, forcing agents to infer spatial context through communication with neighbors in order to coordinate morphogenesis.

Taking inspiration from developmental biology, where most morphogenesis begins from a nearly spherical aggregate of cells, we initialize the population as an approximately uniform sphere composed of a Fibonacci lattice shell and a Poisson-disk sampled core. Onto this sphere, we impose a minimal patterning cue: a polar region consisting of \(N_\text{org}\) nearest-neighbor cells on the shell is designated as an organizer region and is made to uniformly express a single gene that is off in all other cells. We motivate this setup by the localized expression of morphogens at a pole \citep{wartlick2009morphogen, briscoe2015morphogen} or the classical Spemann organizer \citep{harland1997formation, de2006spemann}, which establishes the body axis in early development. We then elongate the sphere by 10\% along the axis passing through this organizer. When introducing additional organizer groups, they are placed \(90^\circ\) apart from the previous axis (with the third axis determined by the right-hand rule), and each group is assigned a different gene that is expressed uniformly within that organizer and off elsewhere. This procedure yields one to three discrete agent types—shown as colored regions in the top of Figure~\hyperref[main:fig4]{3a}—that seed the patterning process. At each training step, we further perturb the initial positions with zero-mean Gaussian noise of magnitude \(0.03\text{--}0.05\) to prevent the model from exploiting frozen disorder in the core of the sphere. We evolve this initial structure to \(t{=}1\) using the SE(3)-equivariant force model in Section~\ref{section:model}. Further details of the initial structure preparation, model architecture, and training procedures are provided in Appendix~\ref{method:agent_based_simulation}.

Following training, we evaluate the learned force model under stronger perturbations by increasing the noise magnitude up to 0.2 (Figure~\hyperref[main:fig4]{3a}, middle) and by shifting the positions of organizer regions while keeping the point cloud fixed (Figure~\hyperref[main:fig4]{3a}, bottom, shown for the noise magnitude 0.05). Since Poisson-disk sampling introduces spatial nonuniformity in the core, shifting the organizers requires spatial cues from the shell to be propagated through agents arranged differently in space. As noise increases, the initial point clouds become highly perturbed, with some cells bulging outward; nevertheless, DiffeoMorph consistently recovers the target geometry, with the final orientation determined by the organizer locations. These results indicate that the learned force model captures a robust, organizer-controlled morphogenetic protocol that generalizes across diverse initial configurations. 

We find that introducing additional organizer groups systematically accounts for reductions in target shape symmetry. Transitioning from an axially symmetric ellipsoid to a starfish with dihedral symmetry requires only one additional organizer (blue) to break rotational symmetry about the axis, rather than assigning a dedicated organizer to each arm. Introducing a third organizer enables the model to capture chirality. As shown in Figure~\hyperref[figure:two_organizers]{S6}, a helical ellipsoid with left-handed chirality can be learned with two organizers; however, under rotation of the organizers, the model fails to preserve the correct handedness. This indicates that reflection symmetry is broken by exploiting noise-induced positional disorder in the training configuration, rather than being encoded through the organizers. Consequently, at least three organizer groups are required to reliably break reflection symmetry and fix chirality. With three organizer groups, the model can define an oriented coordinate frame in 3D, providing sufficient information to generate complex shapes such as the Stanford bunny.

\paragraph{Extension of morphogenesis training.}
In Figure~\hyperref[main:fig4]{3b}, we learn a model for the dynamical behavior of the agents by applying the shape-matching loss at multiple time points. The model is trained to alternate between short and long ellipsoids at intervals of 0.25 over two periods, beginning at \(t{=}0.5\), together with the velocity consistency loss of Equation~\eqref{velocity_consistency_loss}. When integrated beyond the training window, the shape continues to oscillate between these two targets, showing that the learned dynamics persist. In Figure~\hyperref[main:fig4]{3c}, we train the model to retain its final shape after \(t{=}1\), the time at which training constraints terminate, by adding a velocity penalty at \(t{=}1\). The velocity norm converges to zero, and the evolved shape maintains the bunny morphology beyond the training horizon. In Figure~\ref{figure:director_learning}, we further show that the framework can learn shapes with vector-valued weights.

\paragraph{Model benchmarking.}
\begin{table}[t!]
\centering
\resizebox{1\textwidth}{!}{
\begin{tabular}{cc c c c c c c c c c c}
\toprule
 &  & \multicolumn{2}{c}{\textbf{Ellipsoid}} & \multicolumn{2}{c}{\textbf{Crescent}} & \multicolumn{2}{c}{\textbf{Starfish}} & \multicolumn{2}{c}{\textbf{Helical Ellipsoid}} & \multicolumn{2}{c}{\textbf{Bunny}} \\
\cmidrule(lr){3-4} \cmidrule(lr){5-6} \cmidrule(lr){7-8} \cmidrule(lr){9-10} \cmidrule(lr){11-12}
 &  & \textbf{Original} & \textbf{Rotated} & \textbf{Original} & \textbf{Rotated} & \textbf{Original} & \textbf{Rotated} & \textbf{Original} & \textbf{Rotated} & \textbf{Original} & \textbf{Rotated} \\
\midrule
\textbf{GNCA} & No noise &
    \multicolumn{1}{|c}{16.21} & \multicolumn{1}{c|}{17.34{\footnotesize $\pm$0.48}} 
  & 17.32 & \multicolumn{1}{c|}{19.74{\footnotesize $\pm$1.26}} 
  & 8.98 & \multicolumn{1}{c|}{10.43{\footnotesize $\pm$0.70}} 
  & 19.69 &  \multicolumn{1}{c|}{21.43{\footnotesize $\pm$1.39}} 
  & 12.15 & \multicolumn{1}{c}{12.66{\footnotesize $\pm$0.27}} \\
\midrule
\multirow{4}{*}{\textbf{NPA}} 
    & No noise         
        & \multicolumn{1}{|c}{0.19} & \multicolumn{1}{c|}{1.21{\footnotesize $\pm$0.39}} 
      & 0.18 & \multicolumn{1}{c|}{2.44{\footnotesize $\pm$0.97}} 
      & 0.24 & \multicolumn{1}{c|}{5.22{\footnotesize $\pm$2.03}} 
      & 0.15 & \multicolumn{1}{c|}{3.11{\footnotesize $\pm$0.60}}  
      & 0.17 & \multicolumn{1}{c}{5.93{\footnotesize $\pm$3.33}} \\
    & +0.05$\epsilon$  
        & \multicolumn{1}{|c}{0.25{\footnotesize $\pm$0.03}} & \multicolumn{1}{c|}{1.33{\footnotesize $\pm$0.48}} 
      & 0.20{\footnotesize $\pm$0.02} & \multicolumn{1}{c|}{2.57{\footnotesize $\pm$0.98}} 
      & 0.61{\footnotesize $\pm$0.07} & \multicolumn{1}{c|}{5.85{\footnotesize $\pm$2.87}}
      & 0.41{\footnotesize $\pm$0.06} & \multicolumn{1}{c|}{3.23{\footnotesize $\pm$0.58}}  
      & 0.38{\footnotesize $\pm$0.02} & \multicolumn{1}{c}{6.39{\footnotesize $\pm$3.34}} \\
    & +0.1$\epsilon$   
        & \multicolumn{1}{|c}{0.53{\footnotesize $\pm$0.07}} & \multicolumn{1}{c|}{1.76{\footnotesize $\pm$0.55}}
      & 0.53{\footnotesize $\pm$0.09} & \multicolumn{1}{c|}{3.11{\footnotesize $\pm$1.12}}
      & 1.67{\footnotesize $\pm$0.28} & \multicolumn{1}{c|}{6.77{\footnotesize $\pm$1.78}}
      & 1.00{\footnotesize $\pm$0.07} & \multicolumn{1}{c|}{3.82{\footnotesize $\pm$0.63}}   
      & 0.94{\footnotesize $\pm$0.11} & \multicolumn{1}{c}{6.46{\footnotesize $\pm$3.05}} \\
    & +0.2$\epsilon$   
        & \multicolumn{1}{|c}{3.55{\footnotesize $\pm$3.14}} & \multicolumn{1}{c|}{3.39{\footnotesize $\pm$0.97}}
      & 2.05{\footnotesize $\pm$0.21} & \multicolumn{1}{c|}{4.93{\footnotesize $\pm$1.29}}
      & 2.90{\footnotesize $\pm$0.57} & \multicolumn{1}{c|}{8.02{\footnotesize $\pm$1.39}}
      &  3.31{\footnotesize $\pm$0.52} & \multicolumn{1}{c|}{5.47{\footnotesize $\pm$0.75}}   
      & 2.45{\footnotesize $\pm$0.47} & \multicolumn{1}{c}{6.87{\footnotesize $\pm$2.44}} \\
\midrule
\multirow{4}{*}{\textbf{DiffeoMorph (ours)}} 
    & No noise         
        & \multicolumn{1}{|c}{0.38} & \multicolumn{1}{c|}{0.42{\footnotesize $\pm$0.02}} 
      & 0.28 & \multicolumn{1}{c|}{0.35{\footnotesize $\pm$0.03}} 
      & 0.35 & \multicolumn{1}{c|}{0.43{\footnotesize $\pm$0.02}} 
      &  0.24 & \multicolumn{1}{c|}{0.56{\footnotesize $\pm$0.09}} 
      & 0.23 & \multicolumn{1}{c}{0.46{\footnotesize $\pm$0.07}} \\
    & +0.05$\epsilon$  
        & \multicolumn{1}{|c}{0.41{\footnotesize $\pm$0.02}} & \multicolumn{1}{c|}{0.43{\footnotesize $\pm$0.01}} 
      & 0.32{\footnotesize $\pm$0.02} & \multicolumn{1}{c|}{0.42{\footnotesize $\pm$0.05}} 
      & 0.39{\footnotesize $\pm$0.03} & \multicolumn{1}{c|}{0.48{\footnotesize $\pm$0.03}} 
      & 0.51{\footnotesize $\pm$0.08} & \multicolumn{1}{c|}{0.67{\footnotesize $\pm$0.13}}  
      & 0.40{\footnotesize $\pm$0.03} & \multicolumn{1}{c}{0.73{\footnotesize $\pm$0.08}} \\
    & +0.1$\epsilon$   
        & \multicolumn{1}{|c}{0.45{\footnotesize $\pm$0.03}} & \multicolumn{1}{c|}{0.46{\footnotesize $\pm$0.03}}
      & 0.44{\footnotesize $\pm$0.02} & \multicolumn{1}{c|}{0.60{\footnotesize $\pm$0.12}}
      & 0.53{\footnotesize $\pm$0.04} & \multicolumn{1}{c|}{0.75{\footnotesize $\pm$0.09}}  
      & 1.06{\footnotesize $\pm$0.24} & \multicolumn{1}{c|}{1.14{\footnotesize $\pm$0.27}}  
      & 0.78{\footnotesize $\pm$0.06} & \multicolumn{1}{c}{1.02{\footnotesize $\pm$0.13}} \\
    & +0.2$\epsilon$   
        & \multicolumn{1}{|c}{0.51{\footnotesize $\pm$0.04}} & \multicolumn{1}{c|}{0.50{\footnotesize $\pm$0.04}}
      & 0.86{\footnotesize $\pm$0.06} & \multicolumn{1}{c|}{1.08{\footnotesize $\pm$0.33}}
      & 0.80{\footnotesize $\pm$0.06} & \multicolumn{1}{c|}{0.96{\footnotesize $\pm$0.14}} 
      &  2.00{\footnotesize $\pm$0.62}& \multicolumn{1}{c|}{3.31{\footnotesize $\pm$1.10}}  
      & 1.79{\footnotesize $\pm$0.12} & \multicolumn{1}{c}{2.15{\footnotesize $\pm$0.33}} \\

\bottomrule
\end{tabular}
}
\caption{\textbf{Benchmarking against Neural Cellular Automata (NCA) models.} Graph NCA (GNCA) model fails to train effectively, leading to poor test performance. Neural Particle Automata (NPA) achieves the lowest losses on the unperturbed training configurations, but its errors increase substantially when the organizers are rotated. DiffeoMorph performs better in many noisy conditions and achieves lower losses under organizer rotation across all five shapes. All values are evaluated using our spectral loss in units of \(10^{-2}\).}
\label{table:model_benchmarking}
\end{table}

In Table~\ref{table:model_benchmarking}, we benchmark our force model against Graph Neural Cellular Automata (GNCA) and Neural Particle Automata (NPA), both of which support continuous-space agent motion. For a fair comparison, we use each architecture as the force model within our numerical integration scheme and train all models with the same shape-matching loss under the experimental setup described in Section~\ref{method:agent_based_simulation}. The ``Original'' columns report the mean \(\pm\) standard deviation over 10 noise realizations with the organizers in their training orientation. The ``Rotated'' columns additionally average over 10 randomly rotated organizer configurations while keeping the point cloud fixed. Unlike the original GNCA training setup, which provides informative initial features such as normalized target positions or fixed random features, our setup requires models to guide agents using their internal states and minimal spatial cues. This makes the learning task substantially more challenging, and GNCA does not train successfully under these conditions. NPA trains successfully and generalizes well to increasing noise, but its performance deteriorates when the organizers are rotated, consistent with its lack of rotational equivariance. In contrast, our SE(3)-equivariant model achieves comparable generalization to noise while also generalizing across organizer orientations, with losses under organizer rotations remaining close to those obtained with noise alone.

\paragraph{Analysis of the morphogenesis process.}
\begin{figure}[t!]
    \centering

    \includegraphics[width=1\linewidth]{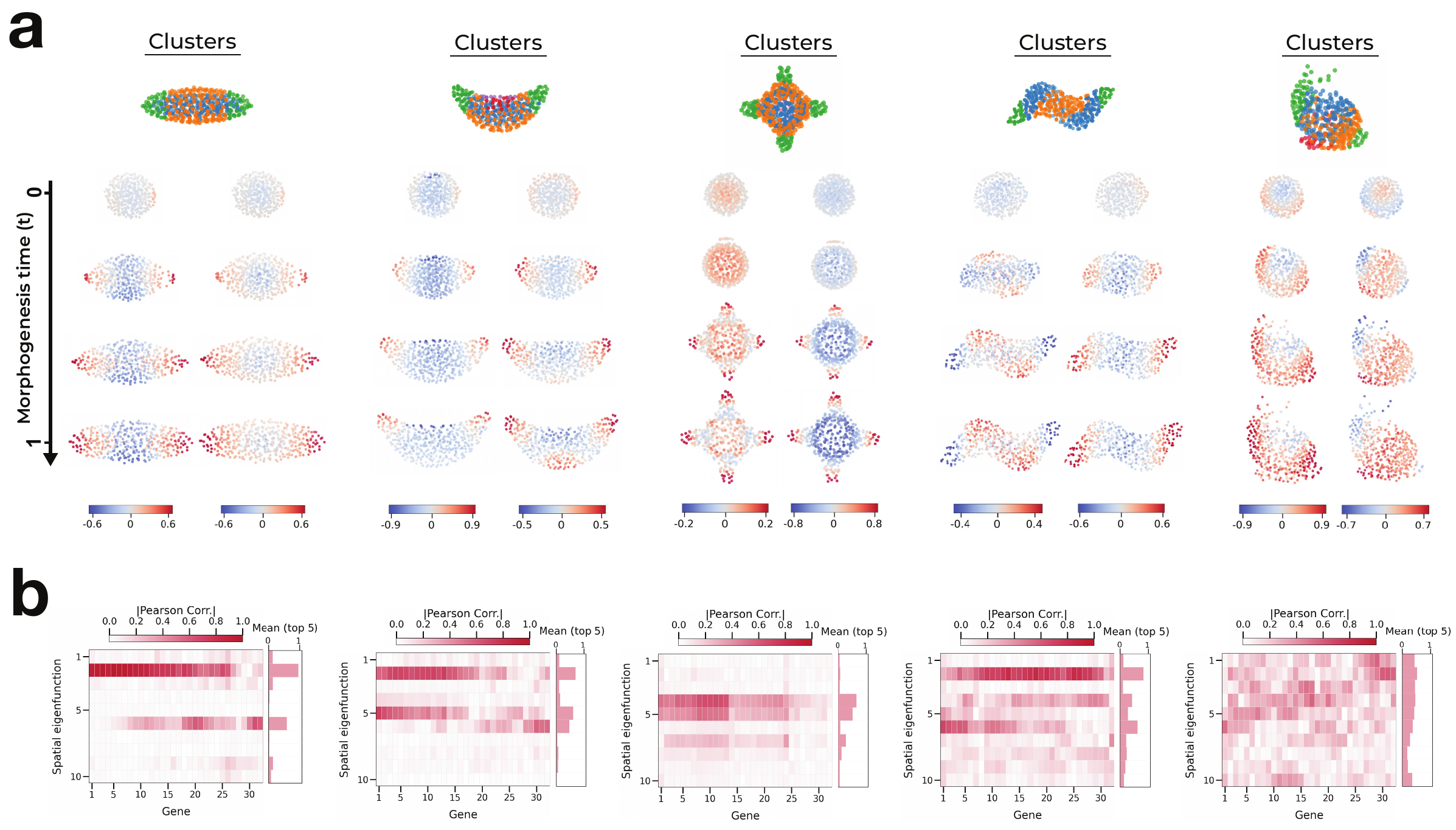}
    
    \caption{\textbf{Analysis of learned gene expression during morphogenesis.} 
(a) Clustering final-time gene expression reveals spatial domains that align with distinct morphological regions. These domains emerge as initially uniform gene expression patterns become spatially structured over time. 
(b) For symmetric shapes, learned gene expression patterns align with a small number of Laplace--Beltrami spatial eigenmodes, whereas for asymmetric shapes such as the bunny, they are distributed more diffusely across modes.}
\label{main:fig5}

\end{figure}

To understand how the model produces robust morphogenesis, we analyze the learned gene expression patterns. In Figure~\hyperref[main:fig5]{4a}, clusters identified from final-time gene expressions, when mapped to spatial coordinates, align well with distinct morphological regions. At the level of individual genes, initially uniform expression evolves into structured spatial patterns—for example, high- and low-expression regions separating poles from the midsection in the ellipsoid, or distinguishing arms from the core in the starfish. For the bunny, the patterns become less symmetric and instead delineate localized morphological compartments. In Figure~\hyperref[main:fig5]{4b}, we compute the Pearson correlation across agents between each learned gene expression pattern and each Laplace--Beltrami eigenmode of the final shape. The learned gene patterns selectively align with a small subset of spatial eigenmodes, as indicated by narrow bands of high correlation. For symmetric shapes such as the ellipsoid, modes inconsistent with symmetry (e.g., antisymmetric left–right modes) are suppressed, and higher-order symmetric modes are utilized instead. In contrast, for the bunny, which lacks strong global symmetry, gene expressions exhibit weaker and more distributed correlations across many modes. Figure~\ref{figure:npa_analysis} shows the same analyses for the non-equivariant NPA model, whose learned gene expression patterns form irregular patches and exhibit more diffuse correlations with the spatial modes. In Table~\ref{table:structure_comparison}, we quantify these correlation patterns using entropy and observe that DiffeoMorph shows lower entropy than NPA, indicating more selective alignment with intrinsic spatial modes. Together, these results suggest that DiffeoMorph learns a parsimonious intrinsic coordinate system through gene expression patterns to represent shape geometry.

\section{Conclusion}
We introduced DiffeoMorph, a differentiable framework for learning how identical agents can self-organize into target 3D shapes. The method combines an SE(3)-equivariant force model with a 3D Zernike shape-matching loss that is invariant to permutation, point count, and rotation while remaining sensitive to reflection. Benchmarks show that the loss has the desired invariance properties and favorable computational scaling. Across target shapes of increasing complexity, DiffeoMorph learns robust morphogenesis from minimal organizer cues, revealing how symmetry-breaking cues and learned internal states can define intrinsic coordinates for shape formation. Future work will explore multi-target training with a single conditional policy and investigate how the number of genes constrains morphogenetic capacity, determining the range and complexity of shapes the agents can form.

\begin{ack}
We thank Christoph Budjan for helpful discussions. This work was supported by the National Heart, Lung, and Blood Institute (NHLBI) of the National Institutes of Health (NIH) under award R01HL158269 and by the Barry Family HSCI Innovation Award for Early Investigators from the Harvard Stem Cell Institute.
\end{ack}


\newpage

\bibliographystyle{plainnat}
\bibliography{reference}

@article{schonemann1966generalized,
  title={A generalized solution of the orthogonal procrustes problem},
  author={Sch{\"o}nemann, Peter H},
  journal={Psychometrika},
  volume={31},
  number={1},
  pages={1--10},
  year={1966},
  publisher={Springer-Verlag}
}

@book{gower2004procrustes,
  title={Procrustes problems},
  author={Gower, John C and Dijksterhuis, Garmt B},
  volume={30},
  year={2004},
  publisher={Oxford University Press, USA}
}

@inproceedings{besl1992method,
  title={Method for registration of 3-D shapes},
  author={Besl, Paul J and McKay, Neil D},
  booktitle={Sensor fusion IV: control paradigms and data structures},
  volume={1611},
  pages={586--606},
  year={1992},
  organization={Spie}
}

@inproceedings{wang2019deep,
  title={Deep closest point: Learning representations for point cloud registration},
  author={Wang, Yue and Solomon, Justin M},
  booktitle={Proceedings of the IEEE/CVF international conference on computer vision},
  pages={3523--3532},
  year={2019}
}

@article{cuturi2013sinkhorn,
  title={Sinkhorn distances: Lightspeed computation of optimal transport},
  author={Cuturi, Marco},
  journal={Advances in neural information processing systems},
  volume={26},
  year={2013}
}

@inproceedings{fan2017point,
  title={A point set generation network for 3d object reconstruction from a single image},
  author={Fan, Haoqiang and Su, Hao and Guibas, Leonidas J},
  booktitle={Proceedings of the IEEE conference on computer vision and pattern recognition},
  pages={605--613},
  year={2017}
}

@article{rubner2000earth,
  title={The earth mover's distance as a metric for image retrieval},
  author={Rubner, Yossi and Tomasi, Carlo and Guibas, Leonidas J},
  journal={International journal of computer vision},
  volume={40},
  pages={99--121},
  year={2000},
  publisher={Springer}
}

@article{memoli2011gromov,
  title={Gromov--Wasserstein distances and the metric approach to object matching},
  author={M{\'e}moli, Facundo},
  journal={Foundations of computational mathematics},
  volume={11},
  pages={417--487},
  year={2011},
  publisher={Springer}
}

@article{peyre2019computational,
  title={Computational optimal transport: With applications to data science},
  author={Peyr{\'e}, Gabriel and Cuturi, Marco and others},
  journal={Foundations and Trends{\textregistered} in Machine Learning},
  volume={11},
  number={5-6},
  pages={355--607},
  year={2019},
  publisher={Now Publishers, Inc.}
}

@inproceedings{mescheder2019occupancy,
  title={Occupancy networks: Learning 3d reconstruction in function space},
  author={Mescheder, Lars and Oechsle, Michael and Niemeyer, Michael and Nowozin, Sebastian and Geiger, Andreas},
  booktitle={Proceedings of the IEEE/CVF conference on computer vision and pattern recognition},
  pages={4460--4470},
  year={2019}
}

@article{osher1988fronts,
  title={Fronts propagating with curvature-dependent speed: Algorithms based on Hamilton-Jacobi formulations},
  author={Osher, Stanley and Sethian, James A},
  journal={Journal of computational physics},
  volume={79},
  number={1},
  pages={12--49},
  year={1988},
  publisher={Elsevier}
}

@inproceedings{park2019deepsdf,
  title={Deepsdf: Learning continuous signed distance functions for shape representation},
  author={Park, Jeong Joon and Florence, Peter and Straub, Julian and Newcombe, Richard and Lovegrove, Steven},
  booktitle={Proceedings of the IEEE/CVF conference on computer vision and pattern recognition},
  pages={165--174},
  year={2019}
}

@article{von1934beugungstheorie,
  title={Beugungstheorie des schneidenver-fahrens und seiner verbesserten form, der phasenkontrastmethode},
  author={von F, Zernike},
  journal={physica},
  volume={1},
  number={7-12},
  pages={689--704},
  year={1934},
  publisher={Elsevier}
}

@article{teague1980image,
  title={Image analysis via the general theory of moments},
  author={Teague, Michael Reed},
  journal={Journal of the optical society of America},
  volume={70},
  number={8},
  pages={920--930},
  year={1980},
  publisher={Optical Society of America}
}

@article{niu2022zernike,
  title={Zernike polynomials and their applications},
  author={Niu, Kuo and Tian, Chao},
  journal={Journal of Optics},
  volume={24},
  number={12},
  pages={123001},
  year={2022},
  publisher={IOP Publishing}
}

@inproceedings{kazhdan2003rotation,
  title={Rotation invariant spherical harmonic representation of 3 d shape descriptors},
  author={Kazhdan, Michael and Funkhouser, Thomas and Rusinkiewicz, Szymon},
  booktitle={Symposium on geometry processing},
  volume={6},
  pages={156--164},
  year={2003}
}

@inproceedings{novotni20033d,
  title={3D Zernike descriptors for content based shape retrieval},
  author={Novotni, Marcin and Klein, Reinhard},
  booktitle={Proceedings of the eighth ACM symposium on Solid modeling and applications},
  pages={216--225},
  year={2003}
}

@article{hu2001angular,
  title={Angular trispectrum of the cosmic microwave background},
  author={Hu, Wayne},
  journal={Physical Review D},
  volume={64},
  number={8},
  pages={083005},
  year={2001},
  publisher={APS}
}

@article{mordvintsev2020growing,
  title={Growing Neural Cellular Automata},
  author={Mordvintsev, Alexander and Randazzo, Ettore and Niklasson, Eyvind and Levin, Michael},
  journal={Distill},
  volume={5},
  number={2},
  pages={e23},
  year={2020},
  doi={10.23915/distill.00023},
  url={https://distill.pub/2020/growing-ca/}
}

@inproceedings{palm2022variational,
  title={Variational Neural Cellular Automata},
  author={Palm, Rasmus Berg and Gonz{\'a}lez-Duque, Miguel and Sudhakaran, Shyam and Risi, Sebastian},
  booktitle={Proceedings of the 10th International Conference on Learning Representations (ICLR)},
  year={2022},
  url={https://arxiv.org/abs/2201.12360}
}

@inproceedings{zhang2021learning,
  title={Learning to Generate 3D Shapes with Generative Cellular Automata},
  author={Zhang, Dongsu and Choi, Changwoon and Kim, Jeonghwan and Kim, Young Min},
  booktitle={Proceedings of the International Conference on Learning Representations (ICLR)},
  year={2021},
  url={https://openreview.net/forum?id=rABUmU3ulQh}
}

@inproceedings{grattarola2021learning,
  title     = {Learning Graph Cellular Automata},
  author    = {Grattarola, Daniele and Livi, Lorenzo and Alippi, Cesare},
  booktitle = {Advances in Neural Information Processing Systems},
  volume    = {34},
  pages     = {20983--20994},
  year      = {2021},
  url       = {https://proceedings.neurips.cc/paper/2021/file/af87f7cdcda223c41c3f3ef05a3aaeea-Paper.pdf}
}

@inproceedings{lin2020modeling,
  title={Modeling 3d shapes by reinforcement learning},
  author={Lin, Cheng and Fan, Tingxiang and Wang, Wenping and Nie{\ss}ner, Matthias},
  booktitle={Computer Vision--ECCV 2020: 16th European Conference, Glasgow, UK, August 23--28, 2020, Proceedings, Part X 16},
  pages={545--561},
  year={2020},
  organization={Springer}
}

@article{viquerat2021direct,
  title={Direct shape optimization through deep reinforcement learning},
  author={Viquerat, Jonathan and Rabault, Jean and Kuhnle, Alexander and Ghraieb, Hassan and Larcher, Aur{\'e}lien and Hachem, Elie},
  journal={Journal of Computational Physics},
  volume={428},
  pages={110080},
  year={2021},
  publisher={Elsevier}
}

@article{pathak2019learning,
  title={Learning to control self-assembling morphologies: a study of generalization via modularity},
  author={Pathak, Deepak and Lu, Christopher and Darrell, Trevor and Isola, Phillip and Efros, Alexei A},
  journal={Advances in Neural Information Processing Systems},
  volume={32},
  year={2019}
}

@article{novotni2004shape,
  title={Shape retrieval using 3D Zernike descriptors},
  author={Novotni, Marcin and Klein, Reinhard},
  journal={Computer-Aided Design},
  volume={36},
  number={11},
  pages={1047--1062},
  year={2004},
  publisher={Elsevier}
}

@article{gala2024n,
  title={E (n)-equivariant Graph Neural Cellular Automata},
  author={Gala, Gennaro and Grattarola, Daniele and Quaeghebeur, Erik},
  journal={Transactions on Machine Learning Research},
  volume={2024},
  number={04},
  year={2024},
  publisher={OpenReview. net}
}

@article{deshpande2025engineering,
  title={Engineering morphogenesis of cell clusters with differentiable programming},
  author={Deshpande, Ramya and Mottes, Francesco and Vlad, Ariana-Dalia and Brenner, Michael P and Dal Co, Alma},
  journal={Nature Computational Science},
  volume={5},
  number={10},
  pages={875--883},
  year={2025},
  publisher={Nature Publishing Group US New York}
}

@article{ramachandran2017searching,
  title={Searching for activation functions},
  author={Ramachandran, Prajit and Zoph, Barret and Le, Quoc V},
  journal={arXiv preprint arXiv:1710.05941},
  year={2017}
}

@inproceedings{satorras2021n,
  title={E (n) equivariant graph neural networks},
  author={Satorras, V{\i}ctor Garcia and Hoogeboom, Emiel and Welling, Max},
  booktitle={International conference on machine learning},
  pages={9323--9332},
  year={2021},
  organization={PMLR}
}

@incollection{hall2013lie,
  title={Lie groups, Lie algebras, and representations},
  author={Hall, Brian C},
  booktitle={Quantum Theory for Mathematicians},
  pages={333--366},
  year={2013},
  publisher={Springer}
}

@book{fulton2013representation,
  title={Representation theory: a first course},
  author={Fulton, William and Harris, Joe},
  volume={129},
  year={2013},
  publisher={Springer Science \& Business Media}
}

@article{geiger2022e3nn,
  title={e3nn: Euclidean neural networks},
  author={Geiger, Mario and Smidt, Tess},
  journal={arXiv preprint arXiv:2207.09453},
  year={2022}
}

@phdthesis{kidger2021on,
  title = {On Neural Differential Equations},
  author = {Patrick Kidger},
  year = {2021},
  school = {University of Oxford},
}

@inproceedings{bridson2007fast,
  title={Fast Poisson disk sampling in arbitrary dimensions},
  author={Bridson, Robert},
  booktitle={ACM SIGGRAPH 2007 sketches},
  pages={22},
  year={2007},
  organization={ACM}
}

@article{gonzalez2010measurement,
  title={Measurement of areas on a sphere using Fibonacci and latitude–longitude lattices},
  author={Gonz{\'a}lez, {\'A}lvaro},
  journal={Mathematical Geosciences},
  volume={42},
  number={1},
  pages={49--64},
  year={2010},
  publisher={Springer}
}

@book{wolpert2015principles,
  title={Principles of development},
  author={Wolpert, Lewis and Tickle, Cheryll and Arias, Alfonso Martinez},
  year={2015},
  publisher={Oxford University Press, USA}
}

@article{scoccimarro2000bispectrum,
  title={The bispectrum: from theory to observations},
  author={Scoccimarro, Rom{\'a}n},
  journal={The Astrophysical Journal},
  volume={544},
  number={2},
  pages={597},
  year={2000},
  publisher={IOP Publishing}
}

@article{kakarala2012bispectrum,
  title={The bispectrum as a source of phase-sensitive invariants for Fourier descriptors: a group-theoretic approach},
  author={Kakarala, Ramakrishna},
  journal={Journal of Mathematical Imaging and Vision},
  volume={44},
  number={3},
  pages={341--353},
  year={2012},
  publisher={Springer}
}

@article{collis1998higher,
  title={Higher-order spectra: the bispectrum and trispectrum},
  author={Collis, WB and White, PR and Hammond, JK},
  journal={Mechanical systems and signal processing},
  volume={12},
  number={3},
  pages={375--394},
  year={1998},
  publisher={Elsevier}
}

@article{gurdon2001morphogen,
  title={Morphogen gradient interpretation},
  author={Gurdon, John B and Bourillot, P-Y},
  journal={Nature},
  volume={413},
  number={6858},
  pages={797--803},
  year={2001},
  publisher={Nature Publishing Group UK London}
}

@article{takeichi2014dynamic,
  title={Dynamic contacts: rearranging adherens junctions to drive epithelial remodelling},
  author={Takeichi, Masatoshi},
  journal={Nature reviews Molecular cell biology},
  volume={15},
  number={6},
  pages={397--410},
  year={2014},
  publisher={Nature Publishing Group UK London}
}

@article{rubenstein2014programmable,
  title={Programmable self-assembly in a thousand-robot swarm},
  author={Rubenstein, Michael and Cornejo, Alejandro and Nagpal, Radhika},
  journal={Science},
  volume={345},
  number={6198},
  pages={795--799},
  year={2014},
  publisher={American Association for the Advancement of Science}
}

@book{sutton1998reinforcement,
  title={Reinforcement learning: An introduction},
  author={Sutton, Richard S and Barto, Andrew G and others},
  volume={1},
  number={1},
  year={1998},
  publisher={MIT press Cambridge}
}

@article{boland2001neural,
  title={A neural network classifier capable of recognizing the patterns of all major subcellular structures in fluorescence microscope images of HeLa cells},
  author={Boland, Michael V and Murphy, Robert F},
  journal={Bioinformatics},
  volume={17},
  number={12},
  pages={1213--1223},
  year={2001},
  publisher={Oxford University Press}
}

@article{alizadeh2016measuring,
  title={Measuring systematic changes in invasive cancer cell shape using Zernike moments},
  author={Alizadeh, Elaheh and Lyons, Samanthe Merrick and Castle, Jordan Marie and Prasad, Ashok},
  journal={Integrative Biology},
  volume={8},
  number={11},
  pages={1183--1193},
  year={2016},
  publisher={Oxford University Press}
}

@article{noll1976zernike,
  title={Zernike polynomials and atmospheric turbulence},
  author={Noll, Robert J},
  journal={Journal of the Optical Society of America},
  volume={66},
  number={3},
  pages={207--211},
  year={1976},
  publisher={Optical Society of America}
}

@article{capalbo2021three,
  title={The Three Hundred project: quest of clusters of galaxies morphology and dynamical state through Zernike polynomials},
  author={Capalbo, Valentina and De Petris, Marco and De Luca, Federico and Cui, Weiguang and Yepes, Gustavo and Knebe, Alexander and Rasia, Elena},
  journal={Monthly Notices of the Royal Astronomical Society},
  volume={503},
  number={4},
  pages={6155--6169},
  year={2021},
  publisher={Oxford University Press}
}

@book{gilbert2020developmental,
  title={Developmental Biology},
  author={Gilbert, Scott F. and Barresi, Michael J. F.},
  edition={12},
  year={2020},
  publisher={Sinauer Associates, Oxford University Press},
  address={New York, NY}
}

@article{whitesides2002self,
  title={Self-assembly at all scales},
  author={Whitesides, George M and Grzybowski, Bartosz},
  journal={Science},
  volume={295},
  number={5564},
  pages={2418--2421},
  year={2002},
  publisher={American Association for the Advancement of Science}
}

@article{fatehullah2016organoids,
  title={Organoids as an in vitro model of human development and disease},
  author={Fatehullah, Aliya and Tan, Si Hui and Barker, Nick},
  journal={Nature cell biology},
  volume={18},
  number={3},
  pages={246--254},
  year={2016},
  publisher={Nature Publishing Group UK London}
}

@article{wartlick2009morphogen,
  title={Morphogen gradient formation},
  author={Wartlick, Ortrud and Kicheva, Anna and Gonz{\'a}lez-Gait{\'a}n, Marcos},
  journal={Cold Spring Harbor perspectives in biology},
  volume={1},
  number={3},
  pages={a001255},
  year={2009},
  publisher={Cold Spring Harbor Lab}
}

@article{briscoe2015morphogen,
  title={Morphogen rules: design principles of gradient-mediated embryo patterning},
  author={Briscoe, James and Small, Stephen},
  journal={Development},
  volume={142},
  number={23},
  pages={3996--4009},
  year={2015},
  publisher={The Company of Biologists}
}

@article{harland1997formation,
  title={Formation and function of Spemann's organizer},
  author={Harland, Richard and Gerhart, John},
  journal={Annual review of cell and developmental biology},
  volume={13},
  number={1},
  pages={611--667},
  year={1997},
  publisher={Annual Reviews 4139 El Camino Way, PO Box 10139, Palo Alto, CA 94303-0139, USA}
}

@article{de2006spemann,
  title={Spemann's organizer and self-regulation in amphibian embryos},
  author={De Robertis, Edward M},
  journal={Nature reviews Molecular cell biology},
  volume={7},
  number={4},
  pages={296--302},
  year={2006},
  publisher={Nature Publishing Group UK London}
}

@article{kim2026neural,
  title={Neural Particle Automata: Learning Self-Organizing Particle Dynamics},
  author={Kim, Hyunsoo and Pajouheshgar, Ehsan and S{\"u}sstrunk, Sabine and Jakob, Wenzel and Park, Jinah},
  journal={arXiv preprint arXiv:2601.16096},
  year={2026}
}

\newpage

\appendix

\setcounter{figure}{0}
\setcounter{table}{0}
\setcounter{equation}{0}
\setcounter{algorithm}{0}

\counterwithout{equation}{section}

\renewcommand{\thefigure}{S\arabic{figure}}
\renewcommand{\theHfigure}{S\arabic{figure}}  

\renewcommand{\thetable}{S\arabic{table}}
\renewcommand{\theequation}{S\arabic{equation}}
\renewcommand{\thealgorithm}{S\arabic{algorithm}}

\renewcommand{\thesection}{\arabic{section}}
\renewcommand{\thesubsection}{\arabic{section}.\arabic{subsection}}

\section{Mathematical Properties of 3D Zernike Polynomials}
\label{supp_sec:zernike}
In this section, we briefly summarize the key properties of the 3D Zernike polynomials. For a complete discussion, we refer readers to \citet{niu2022zernike} and \citet{ novotni2004shape}. 

The 3D Zernike polynomials \(Z_{n\ell}^m(\mathbf{x}) \) form a complete orthonormal basis 
of the Hilbert space \( L^2\!\left(B_1(0)\right)\) and are constructed as the product of:
\begin{align*}
  \text{Radial polynomial}&: \quad R_{n\ell}(r) = (-1)^n \sqrt2 \sqrt{2n+\ell + \frac{1}{2}+1}r^\ell P_n^{\left(\ell+\frac{1}{2},0\right)}\left(1-2r^2\right)  \\
  \text{Spherical harmonics}&: \quad Y_\ell^m(\theta,\phi)=(-1)^m \sqrt{\frac{2\ell+1}{4\pi}\frac{(\ell - \vert m \vert)!}{(\ell+\vert m \vert)!}} P_\ell^{\vert m \vert} (\cos\theta)e^{im\phi} \\
\end{align*}
where \(P_n^{\left(\alpha, \beta\right)}\) and \( P_l^m\) are the Jacobi and the associated Legendre polynomials and we used the orthonormal normalization scheme.

Any function \(f \!\in \! L^2 \!\left(B_1(0)\right)\) can be expanded in the 3D Zernike polynomial basis with coefficients given by the inner product,
\begin{equation}
c_{n \ell}^m =
    \langle  f, Z_{n\ell}^m \rangle =
    \int_{B_1(0)}f\left(r,\theta,\phi\right) \,R_{nl}\left(r\right)\overline{Y_\ell^m \left(\theta,\phi \right)} \, r^2 \sin\theta \, dr \, d\theta \, d\phi
\label{supp:eq1}
\end{equation}
and, once truncated at suitable frequency cutoffs \(n_\text{max}\) and \(\ell_\text{max}\), \(c_{n\ell}^m\) give a compact representation of a function \(f\).

Note that when a function \(f\) is real-valued, the complex-valued spherical harmonics \(Y_l^m\) can be converted to real-valued forms as,
\begin{equation}
Y_{\ell m}^\text{real} =
\begin{cases}
\frac{i}{\sqrt{2}} \left( Y_\ell^{-|m|} - (-1)^m Y_\ell^{|m|} \right) & \text{if } m < 0 \\
Y_\ell^0 & \text{if } m = 0 \\
\frac{1}{\sqrt{2}} \left( Y_\ell^{-|m|} + (-1)^m Y_\ell^{|m|} \right) & \text{if } m > 0
\end{cases}
\end{equation}

This transformation rule defines the real-to-complex change-of-basis \(Q^\ell\), whose entries satisfy
\begin{equation}
  (Q^\ell)_{m m'} =  
    \begin{cases}
      1 & \text{if } m=0 \text{ and } m'=0 \\
      -\frac{i}{\sqrt{2}} & \text{if } m<0 \text{ and } m' = -|m| \\
      \frac{1}{\sqrt{2}} & \text{if } m<0 \text{ and } m' = |m| \\
      \frac{i(-1)^{|m|}}{\sqrt{2}} & \text{if } m>0 \text{ and } m' = -|m| \\
      \frac{(-1)^{|m|}}{\sqrt{2}} & \text{if } m>0 \text{ and } m' = |m| \\
      0 & \text{otherwise}
    \end{cases}
\label{supp:eq2}
\end{equation}
where \(m\) indexes the complex spherical harmonics and \(m'\) indexes the real spherical harmonics.

Using the real-valued form of the spherical harmonics ensures the resulting Zernike polynomials \(Z_{n\ell m}\) and their moments \(c_{n\ell m}\) are also real-valued. This real formulation is commonly adopted in machine learning applications, as most signals or functions of interest are real-valued, and deep learning libraries offer limited support for complex numbers.

In our work, we model the scaled 3D point cloud by the empirical measure \(\mu \left(\mathbf{x}\right)\!=\!\frac{1}{N}\sum_{i=1}^N\omega_i \delta \left(\mathbf{x}{-}\mathbf{x}_i\right) \) and view it as an approximation of a continuous, square-integrable function \( f \!\in \!L^2 \!\left( B_1(0)\right) \). We then proceed with the summation~\eqref{eq2} of the main text as the discrete analogue of the projection~\eqref{supp:eq1}. Implicit in this operation is that we interpret the measure as a generalized function (i.e., \(\mu \!\in \!\mathcal{D}'\!\left(B_1(0)\right) \!\neq \! L^2\!\left(B_1(0)\right)\)) that is dual to the test functions in \( \mathcal{D} \!\left(B_1(0)\right)\), which in our context are the 3D Zernike polynomials. Therefore, the moment computation in Equation~\eqref{eq2} of the main text is mathematically well-posed in the distributional sense. In practice, we can alternatively interpret \eqref{eq2} as a Monte-Carlo estimate of the continuous integral over the unit ball in \eqref{supp:eq1}.

\section{Constructing Representations of the Lie Group \ensuremath{\mathrm{SU}(2)} for Spectral Rotation}
\label{supp_sec:representation}

The spectral rotation matrix \(D_\text{complex}^\ell\) acts on the vector of complex spherical harmonics,
\[
  \mathbf{Y}_\ell \;=\;
  \bigl(Y_\ell^{-\ell},\,Y_\ell^{-\ell+1},\,\dots,\,Y_\ell^{\;\ell}\bigr)^{\!\top},
\]
which spans a \((2\ell{+}1)\)-dimensional complex vector space. For each degree \(\ell\), a rotation of this vector is represented by the unitary, irreducible \((2\ell{+}1)\!\times\!(2\ell{+}1)\) matrix \(D^\ell \), which is called the Wigner--D matrix. It is most often parameterized by Euler angles \( \boldsymbol{\xi} \!= \! (\alpha,\beta,\gamma) \) which specify successive rotations about the $z$, $y$, and $x$ axes, respectively in ZYX convention. It can also be parameterized by a unit quaternion \(\mathbf{q}\!=\!q_w{+}q_x \mathbf{i} {+} q_y \mathbf{j} {+} q_z  \mathbf{k} \!= \!(q_w, q_x, q_y, q_z)\). Below, we describe the procedure for constructing the Wigner--D matrix from both parameterizations. For further details, we refer interested readers to the standard references \citep{hall2013lie,fulton2013representation}.

\subsection{From Euler angles}
\label{supp_sec:representation:euler}

For the Euler angle parameterization, we first need representations, \(J_x,  J_y,  J_z \!\in\! \mathbb{C}^{\left(2\ell+1\right)\times \left(2\ell+1\right)} \), of the Lie algebra \(\mathfrak{su}(2) \), which forms a basis of rotation about \(x,\,y,\,z\)-axis. To build them for each degree \(\ell\) with azimuthal indices \(m{=}-\ell, -\ell+1,  \dots,  \ell\), we first define the ``ladder" operators that shift the index \(m\) by \(\pm1\) as follows,
\[
  \big[J_+\big]_{mm'} =
      \sqrt{(\ell - m')(\ell + m' + 1)} \;\delta_{m,\,m'+1}
  \quad \quad
  \big[J_-\big]_{mm'} =
      \sqrt{(\ell + m')(\ell - m' + 1)} \;\delta_{m,\,m'-1}
\]
where \(\delta_{ij}\) denotes the Kronecker delta, and construct \(J_x\), \(J_y \), and a diagonal \(J_z\) as,
\[
J_x = \frac{1}{2}(J_+ + J_-) \quad \quad
J_y = \frac{1}{2i}(J_+ - J_-) \quad \quad 
\big[J_z\big]_{m, m'} = m \, \delta_{m, m'}
\]

Given these bases of rotation and the Euler angles \((\alpha,\beta,\gamma)\), we can build the irreducible representations of SU(2), in degree \(\ell\) basis, by computing the following matrix exponential,
\[
R_z^\ell(\alpha) = \exp\left(-i \alpha J_z\right) \quad \quad
R_y^\ell(\beta) = \exp\left(-i \beta J_y\right) \quad \quad
R_x^\ell(\gamma) = \exp\left(-i \gamma J_x\right)
\]
and the Wigner--D matrix as,
\begin{equation}
  \mathcal{D}^\ell(\alpha,\beta,\gamma) = 
  R_z^\ell(\alpha)\,R_y^\ell(\beta)\,R_x^\ell(\gamma)
\label{supp:eq3}
\end{equation}

This matrix describes, in spectral space, the sequence of rotation: yaw about the body's \(z\)-axis by \(\alpha\), followed by pitch about the new \(y\)-axis by \(\beta\), and finally roll about the resulting \(x\)-axis by \(\gamma\).

\subsection{From a unit quaternion}
\label{supp_sec:representation:quaternion}

For the unit quaternion parameterization, we can obtain one particular representation of SU(2) at \( \ell = 1/2\) directly via,
\begin{equation}
  U(\mathbf{q}) = \begin{pmatrix}
    \mathbf{q}_a^\ast & \mathbf{q}_b \\
    -\mathbf{q}_b^\ast & \mathbf{q}_a
  \end{pmatrix}
  \quad \text{where} \quad 
  \mathbf{q}_a = q_w + \mathbf{i} \,q_z 
  \quad \text{and} \quad 
  \mathbf{q}_b = q_y + \mathbf{i} \, q_x 
\label{supp:eq4}
\end{equation}
where \( \mathbf{q}_a\) and \(\mathbf{q}_b\) are the Cayley-Klein parameters. 

Noting that \( U \) acts on the fundamental space \( \mathbb{C}^2\), the irreducible representation of SU(2) at an arbitrary \( \ell \) is found by taking the \(2\ell\)-fold tensor power \( U(\mathbf{q})^{\otimes 2\ell}\) and restricting its action to the totally symmetric subspace \( \text{Sym}^{2\ell}(\mathbb{C}^2) \! \subset \! (\mathbb{C}^2)^{\otimes 2\ell} \). This procedure is algebraically represented as, 
\begin{equation}
\begin{gathered}
\mathcal{D}^{\ell}_{mm'}(\mathbf{q})= 
\sum_{\rho}
A^\ell_{m m'}(\rho)\,
\mathbf{q}_a^{\,\ell + m - \rho}
\bar{\mathbf{q}}_a^{\,\ell - \rho - m'}
\mathbf{q}_b^{\,\rho - m + m'}
\bar{\mathbf{q}}_b^{\,\rho} \\
\text{where }\;
A^\ell_{m m'}(\rho)
=
\sqrt{\frac{(\ell + m')!\,(\ell - m')!}{(\ell + m)!\,(\ell - m)!}}
(-1)^{\rho}
\binom{\ell + m}{\rho}
\binom{\ell - m}{\ell - \rho - m'}
\end{gathered}
\label{supp:eq5}
\end{equation}
where \( \bar{z} \) denotes the complex conjugate of \( z \). This expression defines the Wigner--D matrix in \(\mathbb{C}^{(2\ell+1)\times(2\ell+1)}\). 

We use the quaternion-polynomial representation in \eqref{supp:eq5} because it maps quaternion components directly to the Wigner--D entries as finite polynomials in \(q_a,\bar q_a,q_b,\bar q_b\), avoiding intermediate angle-mediated formulas.


\subsection{Conversion to real-valued matrix}
Since we use the real-valued spherical harmonics \(Y^\text{real}_{\ell m}\) obtained by applying the unitary matrix \(Q\) in \eqref{supp:eq2}, the corresponding real-valued Wigner--D matrix \(D^\ell \) used for the rest of this study is obtained by
\begin{equation}
  D^\ell =
  (Q^\ell)^{\dagger} \mathcal{D}^\ell Q^\ell
  \label{eq:real_valued_wigner}
\end{equation}
for both parameterizations. \(^\dagger\) denotes the conjugate transpose. 

\section{Riemannian gradient ascent on $S^3$ for spectral alignment}
\label{supp_sec:riemannian_gradient}

\subsection{Implementation details}
Because 3D rotation is represented by a unit quaternion, the optimization must remain on the three-dimensional unit sphere \( S^3 {\subset} \mathbb{R}^4 \). Differentiating the spectral overlap \( \mathcal{M} \) with respect to the unit quaternion \( \mathbf{q} \) using automatic differentiation produces the usual Euclidean gradient \( \nabla_{\mathbf{q}} \mathcal{M} {\in} \mathbb{R}^4 \), which does not lie in the tangent space \(T_\mathbf{q}S^3\) of the unit sphere at \(\mathbf{q}\). To obtain a gradient direction that respects the unit-norm constraint, we project the Euclidean gradient onto the tangent space at \( \mathbf{q} \), yielding the Riemannian gradient
\begin{equation}
\nabla_{\mathbf{q}}^{S^3} \mathcal{M}
= 
\left( \mathbf{I} - \mathbf{q}\mathbf{q}^\top \right)
\nabla_{\mathbf{q}} \mathcal{M},
\label{eq7}
\end{equation}
where \( \mathbf{I} {-} \mathbf{q}\mathbf{q}^\top \) is the orthogonal projector, denoted \(\mathbf{P}\) hereafter, onto the tangent space of the sphere at \( \mathbf{q} \). 

We update the unit quaternion with ADAM optimizer along the geodesic of this tangent vector using the exponential map:
\begin{equation}
  \mathbf{q}_{k+1} = \text{Exp}_{\mathbf{q}_k}(\eta \, v) \coloneq \cos(\eta \, \Vert v \Vert ) \mathbf{q}_k + \sin(\eta \, \Vert v \Vert ) \frac{v}{\Vert v \Vert}
\label{eq8}
\end{equation}
where \( \eta \) is the step size and \( v {=} \nabla_\mathbf{q}^{S^3} \mathcal{M} \). These steps ensure that \( \mathbf{q} \) remains on the unit sphere after each gradient update and thus continues to represent a valid rotation throughout optimization.

\subsection{Advantage of unit quaternion over Euler angles}
\label{supp_sec:advantages:gradient}

\begin{figure}[h!]
\centering
 \includegraphics[width=1\linewidth]{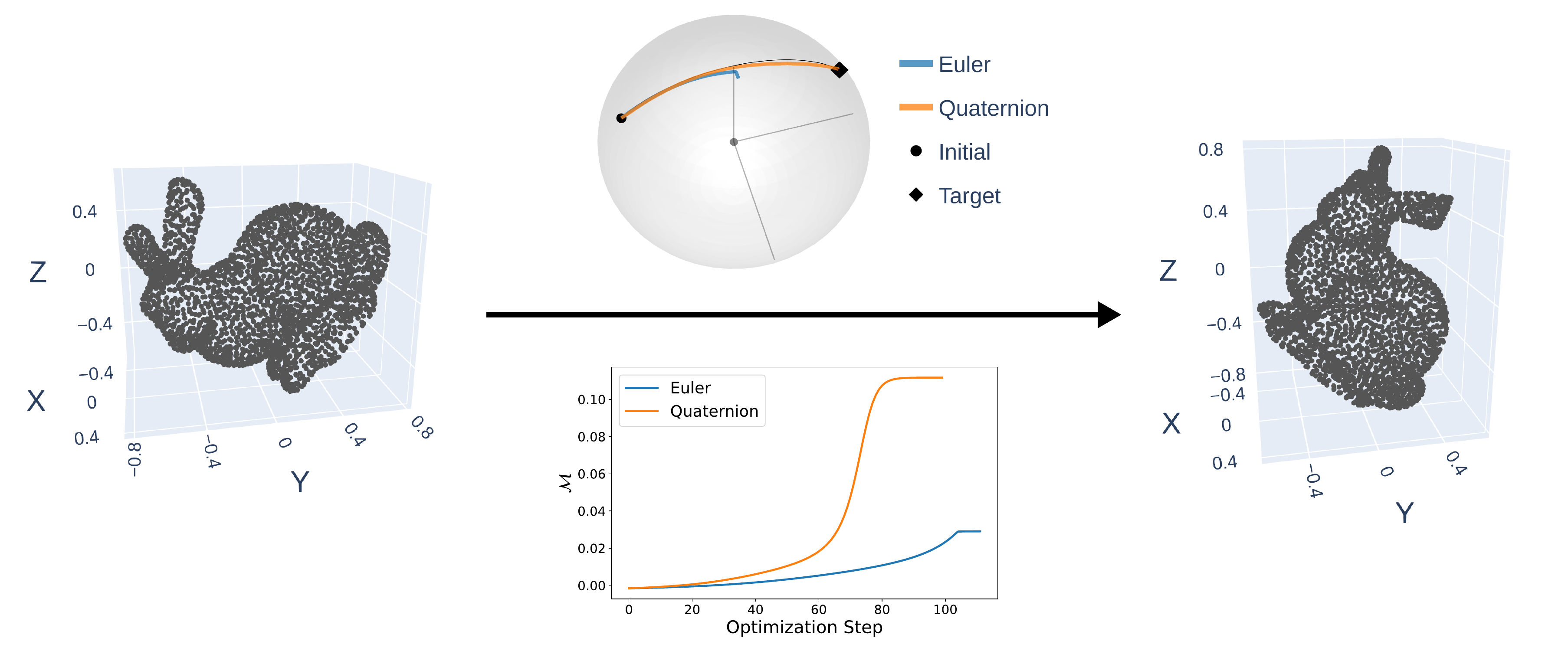}
\caption{\textbf{Comparison of gradient-based optimization using Euler angles and unit quaternion parameterizations.} The goal is to align the spectrum of the bunny rotated by \( +45^\circ \) about the x-axis (left) to that of the bunny rotated by \( -45^\circ \) (right). To visualize the optimization of the two parameterizations both of which have three degrees of freedom on a sphere, we only show the unit rotation axis \( \mathbf{n} \) of the axis-angle coordinates \( (\mathbf{n}, \, \theta)\) where \( \theta \) is the rotation angle about \( \mathbf{n} \) (top center). Note that the alignment can be achieved simply by rotating \( \mathbf{n} \) while keeping \( \theta{=}0^\circ \) throughout. The optimization trajectories of both parameterizations progress smoothly towards the north pole, which is a gimbal lock for Euler angles. At this region, the Euler angle trajectory stalls due to vanishing gradients, whereas the unit quaternion trajectory passes through and reaches the target. Learning curves of the spectral overlap \( \mathcal{M} \) highlight the premature convergence of the alignment optimization when using Euler angles compared to using unit quaternion (bottom center).}
\label{supp:quaternion_alignment}
\end{figure}

Euler angle parameterizations of rotation suffer from gimbal lock, which occurs when two of the three rotation axes become aligned, resulting in a loss of one degree of freedom and a degeneracy in the mapping from Euler angles to physical rotations. Figure~\ref{supp:quaternion_alignment} demonstrates that gimbal lock causes the alignment optimization to prematurely stall as soon as the optimization path reaches a gimbal lock configuration. In this experiment, the task is to align the spectrum of the bunny shown on the left to that of the bunny on the right by maximizing the spectral overlap \( \mathcal{M} \). We compare the efficiency of gradient-based optimizations using Euler angles versus a unit quaternion. The Euler-angle trajectory stalls due to vanishing gradients near the north pole---corresponding to the gimbal-lock condition \( \beta{=}0^\circ \)---and consequently fails to align the two shapes. In contrast, the unit quaternion parameterization avoids this singularity entirely, allowing optimization to proceed smoothly and achieve the correct alignment with maximal spectral overlap.

\section{Implicit differentiation of alignment optimization}
\label{supp_sec:implicit_differentiation}

For an exact solution of the inner alignment problem, the Riemannian gradient of the aligned loss with respect to the unit quaternion vanishes, and the envelope-theorem argument in Section~\ref{section:loss:implicit_differentiation} shows that the alignment optimizer \(\mathbf{q}^\star\) can be detached during the backward pass. The derivation below is only needed for the finite-accuracy case, where the inner optimizer stops at an approximate solution and the residual dependence of the alignment on \(\mathbf{w}\) gives a small correction term. Rather than unrolling all inner optimization steps, we compute this correction by differentiating the local stationarity condition of the spectral-overlap objective on \(S^3\).

\subsection{Derivation of Riemannian Hessian}

Implicit differentiation starts from the local stationarity condition for the inner alignment problem, \(\nabla^{S^3}_\mathbf{q}\mathcal{M}(\mathbf{w},\mathbf{q}^\star(\mathbf{w})){=}0\). Differentiating this condition with respect to \(\mathbf{w}\) gives \(\text{Hess}_{S^3}\mathcal{M}\,\nabla_\mathbf{w}\mathbf{q}^\star+\nabla^2_{\mathbf{q},\mathbf{w}}\mathcal{M}=0\), where \(\text{Hess}_{S^3}\mathcal{M}\) is the Hessian of the spectral overlap restricted to the tangent space of \(S^3\). Solving this linear system for \(\nabla_\mathbf{w}\mathbf{q}^\star\) yields the desired correction term, \(\nabla_\mathbf{w}\mathbf{q}^\star = -\left(\text{Hess}_{S^3}\mathcal{M}\right)^{-1}\nabla^2_{\mathbf{q},\mathbf{w}}\mathcal{M}\), which we now express explicitly in embedded unit-quaternion coordinates.

Taking advantage of the structure of the unit-quaternion optimization problem, we compute the Jacobian term using Riemannian implicit differentiation as follows,
\begin{equation}
  \nabla_\mathbf{w}\mathbf{q}^\star = 
    - \big[\mathbf{P} \, \left( \nabla_{\mathbf{q}\mathbf{q}}^2 \mathcal{M} - \alpha \mathbf{I} \right) \mathbf{P} \big]_{T_\mathbf{q}S^3}^{-1}
    \nabla^2_{\mathbf{q},\mathbf{w}} \mathcal{M},
\label{eq9}
\end{equation}
where \(\nabla^2_{\mathbf{q},\mathbf{w}}\mathcal{M}
= \nabla_\mathbf{w}\circ\nabla^{S^3}_\mathbf{q}\mathcal{M}\) denotes the derivative of the Riemannian quaternion gradient with respect to \(\mathbf w\), expressed as a tangent-vector-valued Jacobian, and \(\alpha=\mathbf{q}^\top\nabla_\mathbf{q}\mathcal{M}\). Here \(\nabla_{\mathbf{q}\mathbf{q}}^2\mathcal{M}\) is the Euclidean Hessian in \(\mathbb{R}^4\), while the projected operator inside the square brackets is the Riemannian Hessian restricted to \(T_\mathbf{q}S^3\).

To derive \eqref{eq9}, we start from the variation of \(\nabla_\mathbf{q}^{S^3} \mathcal{M} \) of \eqref{eq7} under \( \mathbf{q} \!\mapsto \!\mathbf{q} {+} \delta \mathbf{q}\),
\begin{equation}
  \delta \!\left[ \nabla_\mathbf{q}^{S^3} \mathcal{M}\right] 
    = -\left(\delta\mathbf{q} \mathbf{q}^\top + \mathbf{q}\delta\mathbf{q}^\top \right) \nabla_\mathbf{q}\mathcal{M}   + \left(\mathbf{I} - \mathbf{q}\mathbf{q}^\top\right) \nabla_{\mathbf{q}\mathbf{q}}^2 \mathcal{M} \delta \mathbf{q}.
\end{equation}

Once we project it onto the tangent space using \( \mathbf{P}\),
\begin{equation}
\begin{aligned}
 \mathbf{P} \delta \!\left[ \nabla_\mathbf{q}^{S^3} \mathcal{M}\right] 
   &= -\mathbf{P}\delta\mathbf{q} \mathbf{q}^\top\nabla_\mathbf{q}\mathcal{M}
      - \mathbf{P}\mathbf{q}\delta\mathbf{q}^\top\nabla_\mathbf{q}\mathcal{M}
      + \mathbf{P}^2 \nabla_{\mathbf{q}\mathbf{q}}^2 \mathcal{M} \delta \mathbf{q} \\
   &= -\delta\mathbf{q} \mathbf{q}^\top\nabla_\mathbf{q}\mathcal{M}
      + \mathbf{P}\nabla_{\mathbf{q}\mathbf{q}}^2 \mathcal{M} \delta \mathbf{q} \\
   &= \left[\mathbf{P}\nabla_{\mathbf{q}\mathbf{q}}^2 \mathcal{M} - \alpha \mathbf{I}\right] \delta \mathbf{q},
\end{aligned}
\end{equation}
where we use \(\mathbf{P}\delta\mathbf{q}=\delta\mathbf{q}\), \(\mathbf{P}\mathbf{q}=0\), and \(\alpha=\mathbf{q}^\top\nabla_\mathbf{q}\mathcal{M}\).

The above expression is derived as a directional derivative along \( \delta\mathbf{q} \!\in \!T_\mathbf{q}S^3 \). To obtain the generic operator expression of the Riemannian Hessian, \( \text{Hess}_{S^3} \mathcal{M}: T_\mathbf{q}S^3 \rightarrow T_\mathbf{q}S^3\), we multiply with \(\mathbf{P}\) from the right to restrict the domain to the tangent space,
\begin{equation*}
  \text{Hess}_{S^3} \mathcal{M} = \mathbf{P}\left( \nabla_{\mathbf{q}\mathbf{q}}^2 \mathcal{M} - \alpha \mathbf{I}\right) \mathbf{P},
\end{equation*}
which recovers the expression of the Riemannian Hessian.

Since the alignment variable is only a unit quaternion, applying the inverse of the tangent-space Hessian is computationally inexpensive. In implementation, we represent the operator as an embedded \(4{\times}4\) projected Hessian and compute its Moore--Penrose pseudoinverse with a threshold of \(0.01\), which automatically removes the radial null component. In practice, the discreteness of the point cloud, the slight elongation of the initial shape, stochastic noise introduced during simulation, and the use of a pseudoinverse together make the Hessian inversion numerically stable. We discuss the effect of rotational symmetry on Hessian conditioning in Section~\ref{supp_sec:rotational_symmetry}.

\subsection{Finite-tolerance interpretation of the implicit correction}
\label{supp_sec:finite_tolerance_implicit}

The implicit-differentiation formula above is exact for the local stationary solution map \(\mathbf{q}^\star(\mathbf{w})\), not for an arbitrary finite inner-optimization trajectory. Nevertheless, it also gives a useful local correction when the inner solver stops near a nondegenerate stationary point. We now make this statement precise.

Work in local tangent coordinates \(\mathbf{z}\in\mathbb{R}^3\) around the optimal unit quaternion, and write the \(q\)-dependent part of the spectral loss as \(\mathcal{L}(\mathbf{w},\mathbf{z})=A(\mathbf{w})-2\mathcal{M}(\mathbf{w},\mathbf{z})\), where \(A(\mathbf{w})\) collects all terms independent of \(\mathbf{z}\). Let \(\mathbf{z}^\star(\mathbf{w})\) be a nondegenerate local maximizer of \(\mathcal{M}\), so that \(\mathbf{g}:=\nabla_\mathbf{z}\mathcal{M}(\mathbf{w},\mathbf{z}^\star)=0\) and \(\mathbf{H}:=\nabla^2_{\mathbf{z}\mathbf{z}}\mathcal{M}(\mathbf{w},\mathbf{z}^\star)\) is nonsingular. Define \(\mathbf{B}:=\nabla^2_{\mathbf{z}\mathbf{w}}\mathcal{M}(\mathbf{w},\mathbf{z}^\star)\). For the exact aligned objective \(F(\mathbf{w})=\mathcal{L}(\mathbf{w},\mathbf{z}^\star(\mathbf{w}))\), the envelope theorem gives \(\nabla_\mathbf{w}F=\partial_\mathbf{w}\mathcal{L}(\mathbf{w},\mathbf{z}^\star)\).

Now suppose the inner optimization stops at \(\hat{\mathbf{z}}=\mathbf{z}^\star+\mathbf{e}\), with small error \(\mathbf{e}\). The detached gradient is
\[
\mathbf{G}_{\rm det}
=
\partial_\mathbf{w}\mathcal{L}(\mathbf{w},\hat{\mathbf{z}})
=
\nabla_\mathbf{w}F
-
2\mathbf{B}^{\top}\mathbf{e}
+
O(\|\mathbf{e}\|^2),
\]
so detaching an imperfect alignment produces a first-order error in the alignment error \(\mathbf{e}\). Implicit differentiation of the stationarity condition \(\nabla_\mathbf{z}\mathcal{M}(\mathbf{w},\mathbf{z}^\star(\mathbf{w}))=0\) gives \(\nabla_\mathbf{w}\mathbf{z}^\star=-\mathbf{H}^{-1}\mathbf{B}\). Evaluating the corresponding chain-rule term at \(\hat{\mathbf{z}}\) gives
\[
\left(\nabla_\mathbf{w}\mathbf{z}^\star\right)^\top
\nabla_\mathbf{z}\mathcal{L}(\mathbf{w},\hat{\mathbf{z}})
=
\left(-\mathbf{H}^{-1}\mathbf{B}\right)^\top
\left[-2\nabla_\mathbf{z}\mathcal{M}(\mathbf{w},\hat{\mathbf{z}})\right].
\]
Since \(\nabla_\mathbf{z}\mathcal{M}(\mathbf{w},\hat{\mathbf{z}})=\mathbf{H}\mathbf{e}+O(\|\mathbf{e}\|^2)\), this correction equals \(2\mathbf{B}^{\top}\mathbf{e}+O(\|\mathbf{e}\|^2)\). Therefore
\[
\mathbf{G}_{\rm det}
+
\left(\nabla_\mathbf{w}\mathbf{z}^\star\right)^\top
\nabla_\mathbf{z}\mathcal{L}(\mathbf{w},\hat{\mathbf{z}})
=
\nabla_\mathbf{w}F
+
O(\|\mathbf{e}\|^2).
\]
Thus, when the inner solve is close to a nondegenerate stationary point, the implicit term cancels the leading-order error made by detaching the approximate alignment. Equivalently, if \(\mathbf{r}:=\nabla_\mathbf{z}\mathcal{M}(\mathbf{w},\hat{\mathbf{z}})\) is the residual inner gradient, then \(\mathbf{H}^{-1}\mathbf{r}\) estimates the local alignment error, and the implicit correction approximates the gradient that would be obtained after one local Newton correction toward the stationary alignment.

This interpretation is local: implicit differentiation does not give the exact derivative of a finite number of inner optimizer steps. The exact gradient of the implemented finite-step alignment trajectory would require unrolled backpropagation through those steps. The implicit correction is therefore best understood as a finite-tolerance correction to the envelope-gradient approximation, valid when the inner alignment residual is small and the tangent-space Hessian is well conditioned.

\subsection{Discretization and perturbation effects on rotational symmetry}
\label{supp_sec:rotational_symmetry}

\begin{figure}[h!]
    \centering
    \includegraphics[width=1\linewidth]{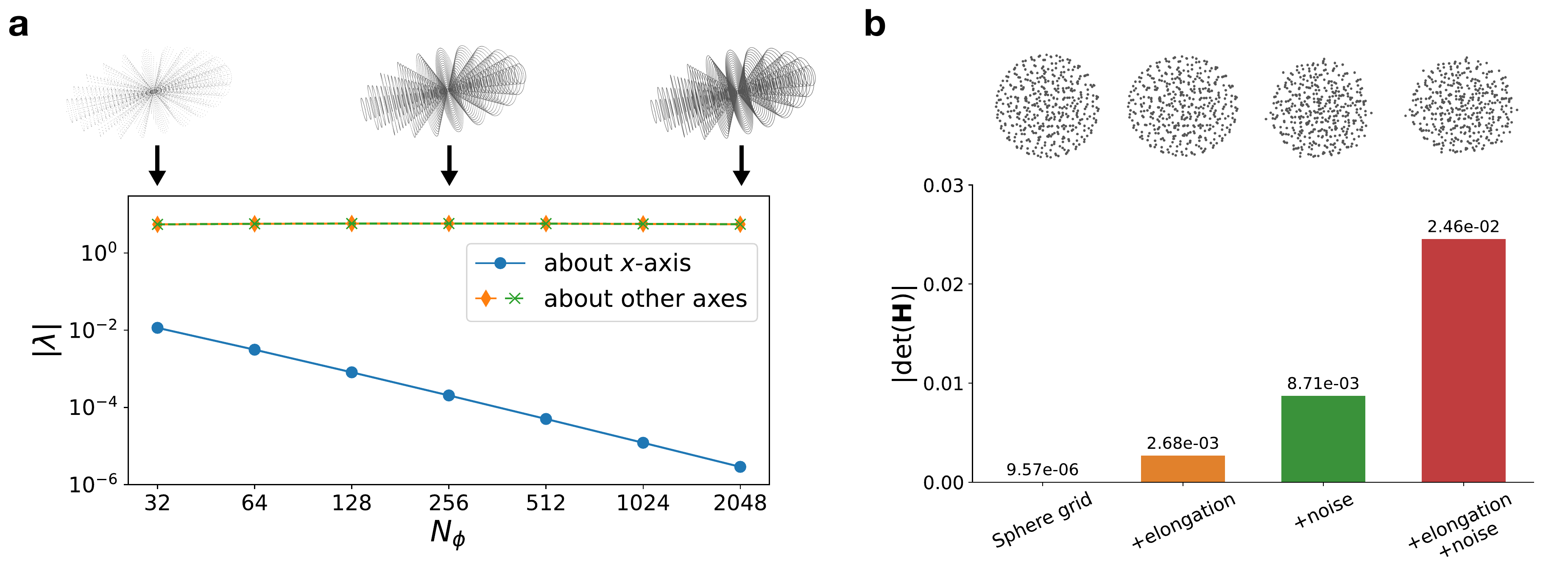}  
    \caption{\textbf{Numerical verification of rotational symmetry.} (a) Magnitudes of the eigenvalues of rotational Hessian \( \mathbf{H}_\mathbf{q} \) of ellipsoid-shaped point clouds with varying number of azimuthal partitions \( N_\phi \) around the symmetry (elongation) axis, labeled as x-axis. One eigenvalue is several orders of magnitude smaller than the other two, reflecting the axisymmetric nature of the ellipsoid. As \( N_\phi \) increases, the \( N_\phi \)-fold cyclic symmetry approaches continuous symmetry, and the eigenvalue along the symmetry axis decreases in line with this trend. (b) Determinant of \( \mathbf{H}_\mathbf{q} \) of sphere-shaped point clouds under different perturbations. The unperturbed sphere grid yields a small but nonzero determinant due to its discrete nature and small nonuniformity induced by the Poisson-disk sampling. Adding noise increases the determinant more than elongation does, since the elongated shape remains symmetric about one axis. Combining elongation and noise produces a shape with a determinant more than three orders of magnitude larger, which is used in the experiments shown in the main text.}
    \label{supp:rotational_symmetry}
\end{figure}

As described above, the Riemannian implicit-differentiation correction requires solving a linear system involving the rotational Hessian, equivalently applying the inverse of this Hessian on the tangent space of the unit-quaternion manifold. The connection between rotational symmetry and Hessian invertibility is easiest to see geometrically. If a shape has an exact continuous rotational symmetry, then there is a nonzero infinitesimal rotation direction in which the shape can be moved without changing its Zernike spectrum, and therefore without changing either the spectral overlap \(\mathcal{M}\) or the aligned shape-matching loss. Along this one-parameter family of rotations, the objective is constant. Hence both the first derivative and the second derivative in that tangent direction vanish. This tangent direction is therefore a null vector of the rotational Hessian, so the Hessian has a zero eigenvalue and cannot be inverted. Conversely, perturbations that break the symmetry lift this flat direction and make the corresponding Hessian eigenvalue nonzero, although it may remain small for nearly symmetric shapes.

Since the morphogenesis simulation operates on a discrete point cloud, it is not immediately clear to what extent the Hessian is sensitive to symmetries that are only approximately represented by discrete points. To examine this, we study how the magnitudes of the Hessian's eigenvalues scale with the number of azimuthal partitions \( N_\phi \) around the symmetry axis of an ellipsoid (Figure~\hyperref[supp:rotational_symmetry]{S2a}). We evaluate the spectral overlap \(\mathcal{M}\) using \(\ \mathbf{C}^\text{evol} {=}\mathbf{C}^\text{target}\), obtained from an \(x\)-axis aligned ellipsoid, and compute its gradient with respect to a unit quaternion that rotates the ellipsoid by \(+1^\circ\) about the \(x\)-axis. This quaternion \(\mathbf{q}_x\) is constructed by converting the axis-angle coordinate \( (\mathbf{n}, \,\theta)\) with \( \mathbf{n}\) coinciding with the \(x\)-axis. The resulting Hessian in \( \mathbb{R}^4 {\times} \mathbb{R}^4 \) is converted into the tangent space Hessian in \( T_{\mathbf{q}_x}S^3 {\times} T_{\mathbf{q}_x}S^3 \) using an orthonormal frame \( \mathbf{U} {\in} \mathbb{R}^{4\times3} \). We obtain this frame via QR decomposition of the matrix whose first column is \(\mathbf{q}_x\) and the remaining columns are chosen from basis vectors in \( \mathbb{R}^4 \). Any potential sign flips in the resulting frame do not affect our analysis, as we compare only the absolute magnitudes of the eigenvalues. We find that the rotational symmetry is already evident in the discrete point cloud at resolutions as low as \( N_\phi {=}32 \), as indicated by several orders of magnitude of separation in one of the tangent-space eigenvalues. As \( N_\phi \) increases and the ellipsoidal point cloud forms smooth rings, one rotational eigenvalue becomes negligibly small, reflecting the discrete \( N_\phi \)-fold cyclic symmetry of the point cloud approaching the true continuous rotational symmetry.

Figure~\hyperref[supp:rotational_symmetry]{S2b} examines the behavior of the Hessian determinant in the actual experimental setting of DiffeoMorph, using the shell+core sphere grid described in Section~\ref{method:initialization:positions} and adding relevant perturbations on it. We compute the spectral overlap \(\mathcal{M}\) using the Zernike moments \(\mathbf{C}^\text{sph}\) of the sphere-derived point clouds as both input and target---i.e., \(\mathcal{M} (\mathbf{C}^{\text{sph}},\mathbf{C}^{\text{sph}} ) \)---and evaluate the Hessians at the unit quaternion corresponding to \(+1^\circ\) rotation about the \(x\)-axis. We verify that the results do not depend on the specific choice of unit quaternion. Because a sphere is symmetric about any axis, the original sphere-shaped point cloud yields a very small Hessian determinant. Elongating the sphere breaks this symmetry and reduces the number of small eigenvalues to one, and the determinant value increases accordingly. Separately, adding zero-mean Gaussian noise with magnitude 0.05 perturbs the grid enough to distort the spherical appearance, leading all three rotational eigenvalues to increase. In DiffeoMorph experiments, these two perturbations are applied jointly: elongation serves as morphogen-mediated symmetry breaking, while noise encourages generalizable learning. The resulting point cloud has a substantially larger determinant than the original sphere grid. This indicates that the gradient of the inner alignment step can be computed robustly via implicit differentiation, which requires inverting the Hessian.

\subsection{Computational cost of inner alignment gradients}
\label{supp_sec:runtime}

\begin{figure}[ht!]
\centering
 \includegraphics[width=0.65\linewidth]{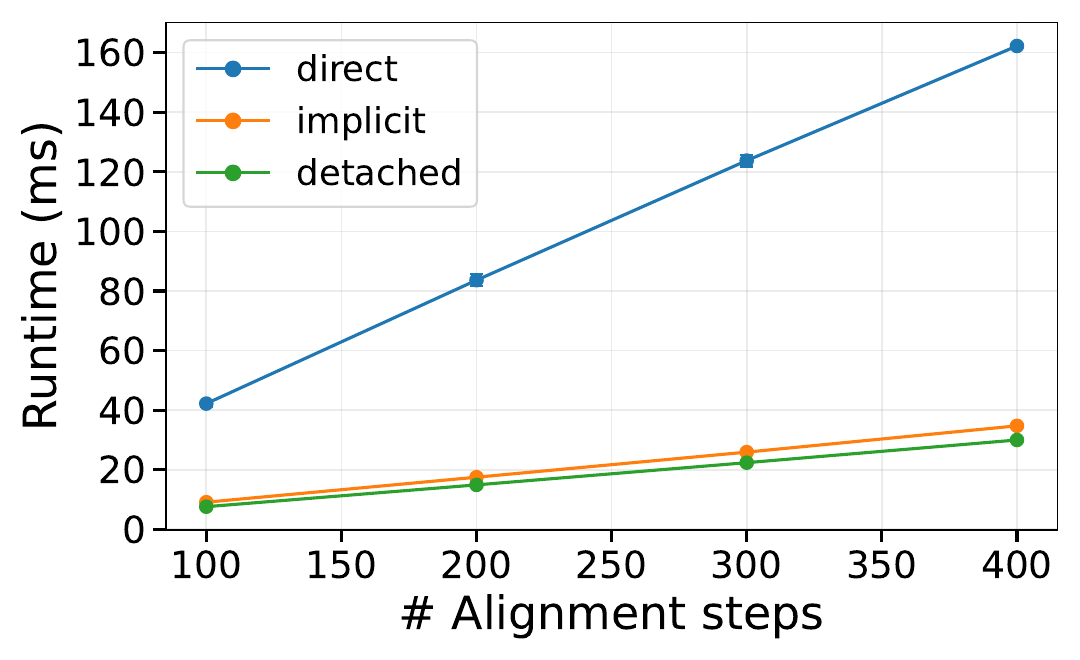}
\caption{\textbf{Runtime comparison for differentiating the shape-matching loss.} Average runtime per direct \(\mathbf{X}\)-optimization step (\(n=5\)) as a function of the number of inner optimization steps used for spectral alignment. Standard unrolled backpropagation (``direct'') is approximately six times slower than detaching the inner alignment gradient (``detached''), while implicit differentiation (``implicit'') has nearly the same runtime as the detached approach.}
\label{supp:runtime}
\end{figure}

To compare the computational cost of computing the gradient through the inner optimization, we perform the direct shape optimization experiment from Figure~\hyperref[main:fig2]{2b} of the main text with \(N\!=\!500\) point cloud using unrolled direct backpropagation, implicit differentiation, and the detached case where we skip computing the inner gradient entirely. We measure the runtime of a single outer training step---which involves a forward step (including alignment steps), a backward pass (where the gradient through alignment steps is computed via backpropagation, implicit differentiation, or skipped entirely in ``direct", ``implicit", and ``detached", respectively), and an outer gradient update---as a function of the number of alignment steps taken for optimizing unit quaternion. After compiling the training function, we execute it on the GPU five times and report the mean values in Figure~\ref{supp:runtime}. We omit showing error bars because the runtime fluctuation is minimal. In all three methods, runtime increases with the number of alignment steps since these iterations correspond directly to the forward computations within the loss. However, direct backpropagation must unroll all forward iterations during the backward pass, whereas implicit differentiation computes the gradient \( \nabla_\mathbf{X}\mathbf{q}^\star \) in a single step. This yields a substantial runtime advantage, as the runtime curve of the implicit method is comparable to that of the detached case.

\section{Details of SDE integration}
\label{supp_sec:pseudocode}

The pseudocode of the computation procedure in our SE(3)-equivariant force model is shown in Algorithm~\ref{pseudocode}. 

\begin{algorithm}[ht!]
\caption{Force computation in the SE(3)-Equivariant Morphogenesis Model}
\begin{algorithmic}
\State \textbf{Input:} Initial positions $\mathbf{X} = \{\mathbf{x}_i\}$, gene expressions $\mathbf{G} = \{\mathbf{g}_i\}$, and noise strengths $\sigma_{\mathbf{x}/\mathbf{g}}$
\State \textbf{Learned model components:}
\Statex \quad $\phi_e$: edge message MLP
\Statex \quad $\phi_x$: force-scaling MLP
\Statex \quad $\phi_g$: gene update MLP

\State \textbf{Initialize:} $t \gets 0$
\While{$t < T$}
    \ForAll{pairs of agents $(i, j)$}
        \State Compute distance: $d_{ij}^2 \gets \|\mathbf{x}_i - \mathbf{x}_j\|^2$
        \State Construct edge feature: $\mathbf{e}_{ij} \gets [\, \mathbf{g}_i \, \| \,\mathbf{g}_j \, \| \, d_{ij}^2 \,]$
        \State Compute message: $\mathbf{m}_{ij} \gets \phi_e(\mathbf{e}_{ij})$
    \EndFor
    \ForAll{agents $i$}
        \State Aggregate message: $\mathbf{m}_i \gets \sum_{j \ne i} \mathbf{m}_{ij}$
        \State Compute velocities: 
        \begin{align*} 
        \text{(Position)} \qquad \mathbf{v}_i &\gets \frac{1}{N-1}\sum_{j \neq i}\phi_{\mathbf{x}}(\mathbf{m}_{ij}) \frac{\mathbf{x}_i - \mathbf{x}_j}{\Vert\mathbf{x}_i - \mathbf{x}_j\Vert}  \\
        \text{(Gene)} \qquad
        \mathbf{u}_i &\gets \phi_\mathbf{g} \left(\mathbf{g}_i \Vert \mathbf{m}_i \right) 
        \end{align*}
    \EndFor
    \ForAll{agents $i$}

        \State Integrate position: $\mathbf{x}_i \gets \mathbf{x}_i + \mathbf{v}_i \Delta t + \sigma_\mathbf{x}\sqrt{\Delta t} \, \boldsymbol{\xi}_i^\mathbf{x}$
        \State Integrate gene: $\mathbf{g}_i \gets \mathbf{g}_i + \mathbf{u}_i  \Delta t + \sigma_\mathbf{g} \sqrt{\Delta t} \, \boldsymbol{\xi}_i^\mathbf{g}$
    \EndFor

    \State Increment time: $t \gets t + \Delta t$
\EndWhile
\end{algorithmic}
\label{pseudocode}
\end{algorithm}

\section{Benchmarking }
\subsection{Loss benchmarking}
\label{supp_sec:loss_benchmarking}

\begin{figure}[ht]
\centering
 \includegraphics[width=1\linewidth]{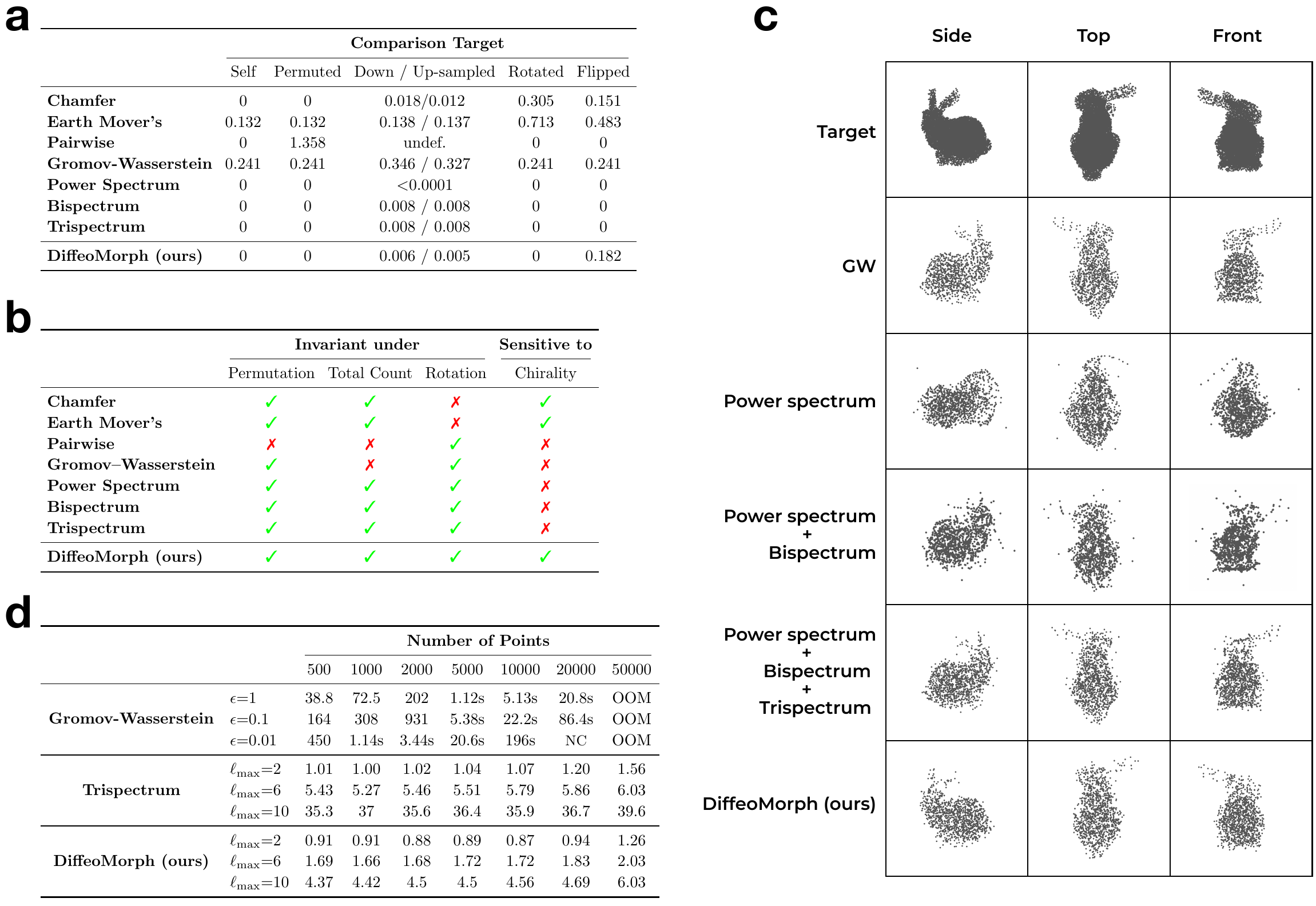}
\caption{\textbf{Benchmarking DiffeoMorph loss.} (\textbf{a}) Distances between the original bunny point cloud (Self) and its geometrically perturbed variants are computed using standard losses for shape comparison. For optimal transport--based losses (Earth Mover's and Gromov--Wasserstein), the distances between identical shapes (Self vs. Self) are nonzero due to the probabilistic relaxation introduced by the entropic regularization, which is required to make them differentiable. (\textbf{b}) Behavior of each loss summarized based on (a). Our loss is the only distance metric satisfying all desired properties. (\textbf{c}) Visualizations of point clouds learned through the direct shape optimization setup of Figure~\hyperref[main:fig2]{2b} using losses satisfying the three invariance properties. We compute higher-order spectra cumulatively: the bispectrum includes the power spectrum, and the trispectrum includes both the power spectrum and bispectrum. The correct head direction is recovered only when training with our loss. (\textbf{d}) Runtime analyses. Gromov--Wasserstein distance scales poorly with the number of points and the weakening of regularization, captured by decreasing \( \epsilon \). In contrast, the runtime of the spectra-based losses is unaffected by the number of points, since the summation over points during projection step is vectorized. However, computing the trispectrum is slower than our loss because it requires enumerating valid quartets of angular degrees \( (\ell_1, \ell_2, \ell_3, \ell_4)\) to capture fourth-order coupling. The spectral alignment step in our loss is faster than this enumeration process, even when vectorized, resulting in the best overall runtime performance.}
\label{figure:loss_benchmarking}
\end{figure}

In this section, we perform several benchmarking experiments to highlight its advantages over standard distance metrics for shape comparison. For comparison, we use Chamfer distance \citep{fan2017point} (nearest-neighbor matching), Earth Mover's distance \citep{rubner2000earth} (optimal transport between point sets), Pairwise distance \citep{gala2024n} (comparison of inter-point distance matrices), Gromov--Wasserstein \citep{memoli2011gromov} distance (optimal transport that matches pairwise distances), Power Spectrum \citep{kazhdan2003rotation, novotni20033d, novotni2004shape} (rotation-invariant magnitudes of Zernike coefficients), bispectrum \citep{scoccimarro2000bispectrum, kakarala2012bispectrum, collis1998higher} (third-order rotation-invariant couplings of angular modes), and trispectrum \citep{collis1998higher} (fourth-order rotation-invariant couplings with higher geometric detail). The mathematical expressions for these distances are provided in Section~\ref{supp_sec:expressions_of_metrics}.

\subsubsection{Geometric properties}

In Figure~\hyperref[figure:loss_benchmarking]{S4a}, we compute the distances between the bunny point cloud and several geometrically perturbed variants. We also include the comparison against itself (Self) to establish a baseline value for each metric. The nonzero Self-to-Self distances for Earth Mover's and Gromov--Wasserstein distances arise from the entropic regularization used to relax the optimal transport problem into a smooth, differentiable objective \citep{cuturi2013sinkhorn, peyre2019computational}. 

Chamfer and Earth Mover's distances compare two point clouds by allowing any point in one cloud to match to any point in the other—via nearest-neighbor matching for Chamfer and via an optimal transport plan for Earth Mover's—rather than relying on a fixed \(i\)-to-\(j\) correspondence. This flexibility makes them invariant to permutations of the points and able to handle point clouds of different sizes. However, they are sensitive to rotation: rotating the target yields a large distance because the comparison is made directly on raw coordinates. Pairwise distance addresses this issue by comparing distances within each point cloud, \( \Vert \mathbf{x}_i {-} \mathbf{x}_j \Vert {-} \Vert \mathbf{y}_i {-} \mathbf{y}_j \Vert  \), which makes it rotation invariant. However, it is sensitive to index permutation and total count because it requires an explicit point-to-point correspondence. 

Gromov--Wasserstein distance combines the strengths of both approaches by finding the optimal probabilistic coupling \( \boldsymbol{\pi}_{ij} \) that minimizes discrepancies between all pairwise distances across two point clouds. As a result, it yields the same value for permuted and rotated point clouds as in the Self comparison. However, changes in the number of points affect the coupling pattern in \( \boldsymbol{\pi}_{ij} \), making the metric sensitive to total point count. 

The spectral metrics avoid these issues entirely. The power spectrum is obtained by summing the squared magnitudes of the 3D Zernike moments over the azimuthal index \(m\), yielding a single energy value for each angular degree \(\ell\). This summation removes the directional information encoded in the \(m\)-components and therefore discards substantial geometric detail. The bispectrum and trispectrum retain more structure by coupling pairs and quartets of angular degrees, respectively, to form rotation-invariant descriptors. Although these higher-order spectra capture finer geometric correlations than the power spectrum, they remain reduced representations of the full \((\ell,m)\)-dependent Zernike moments, and thus lose information relative to them. As a consequence of their construction, the power spectrum, bispectrum, and trispectrum all become insensitive to reflections and therefore cannot distinguish chiral configurations. Only Chamfer and Earth Mover's distances---which compare the raw coordinates of the point clouds---and our loss, which retains the azimuthal components, are sensitive to reflection. These properties are summarized in Figure~\hyperref[figure:loss_benchmarking]{S4b}. 

Among the metrics considered, only our loss simultaneously satisfies all four desired criteria for shape matching: invariance to point permutations, robustness to differing point counts, invariance to rotations, and sensitivity to chirality.

\subsubsection{Direct shape learning}

To understand how these properties affect learning, we performed the direct shape learning task with Gromov--Wasserstein distance (shown as GW) and the spectrum-based metrics. The learned shapes obtained using each metric are rotated to align with the target shape and are visualized together in Figure~\hyperref[figure:loss_benchmarking]{S4c}. Optimization with GW and with all three rotation-invariant spectral metrics can reproduce the target bunny shape, but only up to reflection. Because a reflection-invariant metric cannot distinguish a shape from its mirror image, the optimization has no way to prefer one head orientation over the other. As a result, the learned bunny may emerge facing either direction, depending solely on the initial configuration of the ellipsoid. We also find the power spectrum alone, or even in combination with the bispectrum, lacks expressive power to capture fine geometric details such as the bunny's ears. This suggests that fourth-order coupling through the trispectrum is necessary for accurate shape reconstruction. Our loss, by preserving azimuthal information, successfully recovers the bunny with high fidelity and the correct head orientation. 

\subsubsection{Runtime analysis}

In Figure~\hyperref[figure:loss_benchmarking]{S4d}, we benchmark the evaluation runtime of Gromov--Wasserstein distance, Trispectrum (alone) and our loss. Values are reported in milliseconds unless explicitly marked in seconds. Gromov--Wasserstein scales poorly with increasing point count \(N\) because each iteration requires manipulating full \(N {\times} N\) matrices and repeatedly forming products between them, leading to quadratic memory usage and cubic computational cost. Notably, it reaches a not-converged error (denoted as ``NC") at \(N{=}20000\) and \(\epsilon{=}0.01\). an out-of-memory error (denoted ``OOM") at \( N{=}50000\) because it stores the coupling matrix and two pairwise distance matrices, each of size \( N {\times} N\). 
Lowering the regularization parameter \( \epsilon \) slows the algorithm because the Sinkhorn updates require many more iterations to converge. The trispectrum and our loss do not scale with \( N \) since the point cloud is first projected onto Zernike polynomials and subsequently handled via Zernike moments. Increasing the number of angular components \(\ell_\text{max}\) increases the runtime in both cases, but to different degrees. The trispectrum's \( \ell_\text{max} \)-scaling is due to enumerating quartets of angular degrees \( (\ell_1, \ell_2, \ell_3, \ell_4)\) that satisfy the selection rules. In contrast, DiffeoMorph's runtime grows more slowly because the dominant cost comes from evaluating Wigner--D matrix for each \( \ell \), which can be efficiently vectorized. Therefore, even with the intermediate spectral alignment, our loss achieves the most competitive runtime. Taken together, these results show that our shape-matching loss uniquely combines the geometric invariances, expressive power, and computational efficiency needed for reliable 3D shape learning.

\section{Additional experiments}

\subsection{Spectral alignment and direct shape learning with weighted shapes}
\label{supp_sec:additional:weighted}

\begin{figure}[t!]
    \centering
    \includegraphics[width=0.92\linewidth]{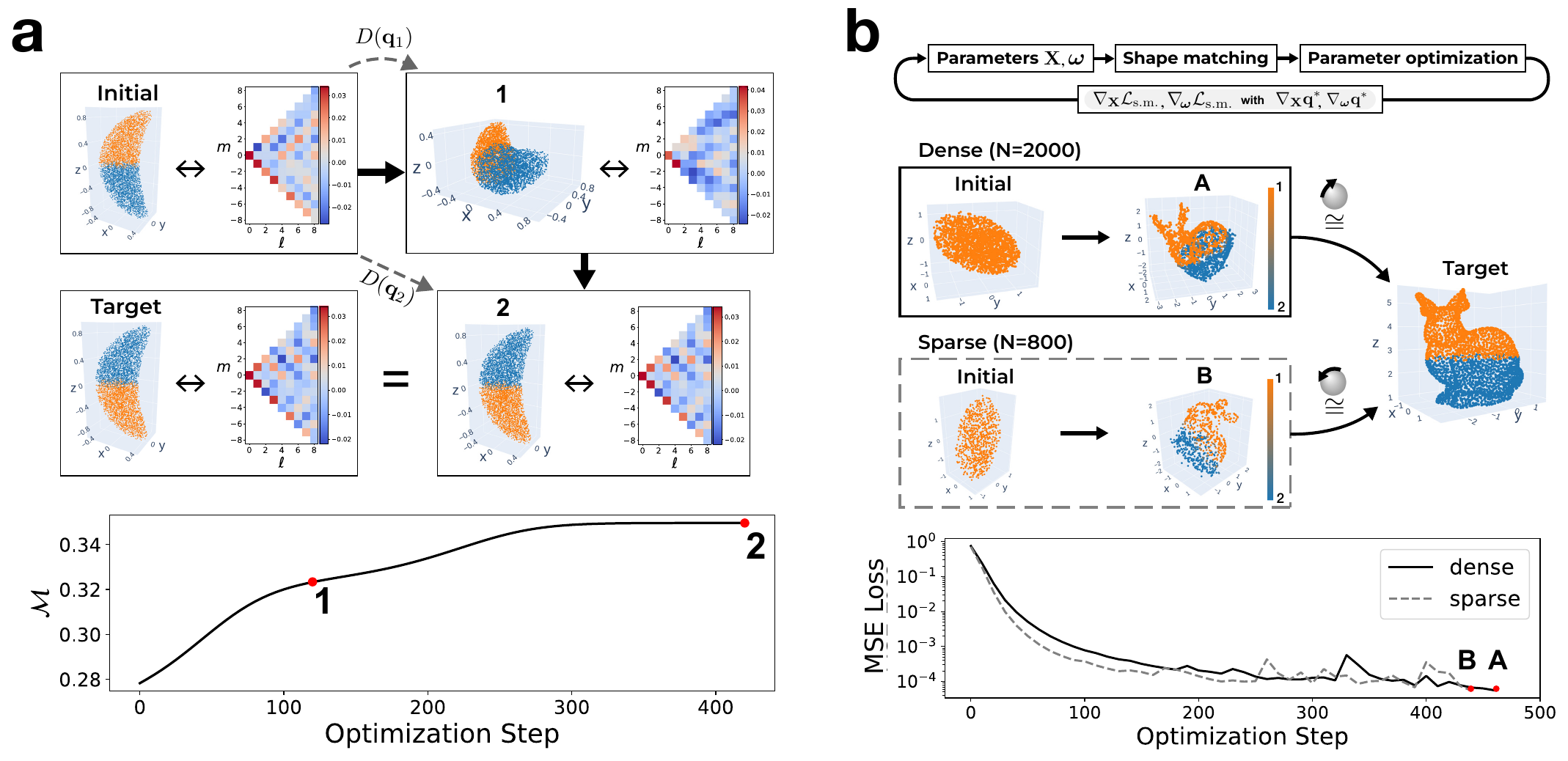}
    \caption{\textbf{Aligning and learning weighted shapes.} The experiments in Figure~\ref{main:fig2} are repeated with weighted point clouds. Orange and blue colors correspond to the weight values of 1 and 2, respectively. (a) Quaternion optimization aligns the weighted crescent to the target by a \(180^\circ\) rotation, matching both shape and weight distribution. (b) Joint optimization of positions \(\mathbf{X}\) and weights \(\boldsymbol{\omega}\) deforms dense and sparse ellipsoids into the target bunny while recovering its spatial weight pattern.} 
    \label{figure:weighted_shape}
\end{figure}

In this section, we repeat the experiments from Figure~\ref{main:fig2} of the main text using shapes associated with weights \(\boldsymbol{\omega}\). Specifically, we consider weighted point clouds in which orange and blue denote weights of 1 and 2, respectively. As shown in Figure~\hyperref[figure:weighted_shape]{S5a}, optimizing over unit quaternions yields a \(180^\circ\) rotation of the crescent, correctly aligning regions with distinct weights to match the target configuration. In Figure~\hyperref[figure:weighted_shape]{S5b}, we jointly optimize the point cloud \(\mathbf{X}\) and weights \(\boldsymbol{\omega}\), starting from ellipsoidal point clouds with initially uniform weights. As learning proceeds, the point clouds deform into the target bunny shape while accurately recovering the spatial weight distribution, assigning weight 2 to the lower half of the body. The global orientation remains unchanged, consistent with the rotation-invariance behavior observed in the unweighted setting. These results show that Zernike moments can faithfully represent weighted shapes and that the weight distribution can be learned jointly with the morphology.

\subsection{Learning chiral shape with two organizers}
\label{supp_sec:additional:two_organizers}

\begin{figure}[t!]
    \centering
    \includegraphics[width=0.9\linewidth]{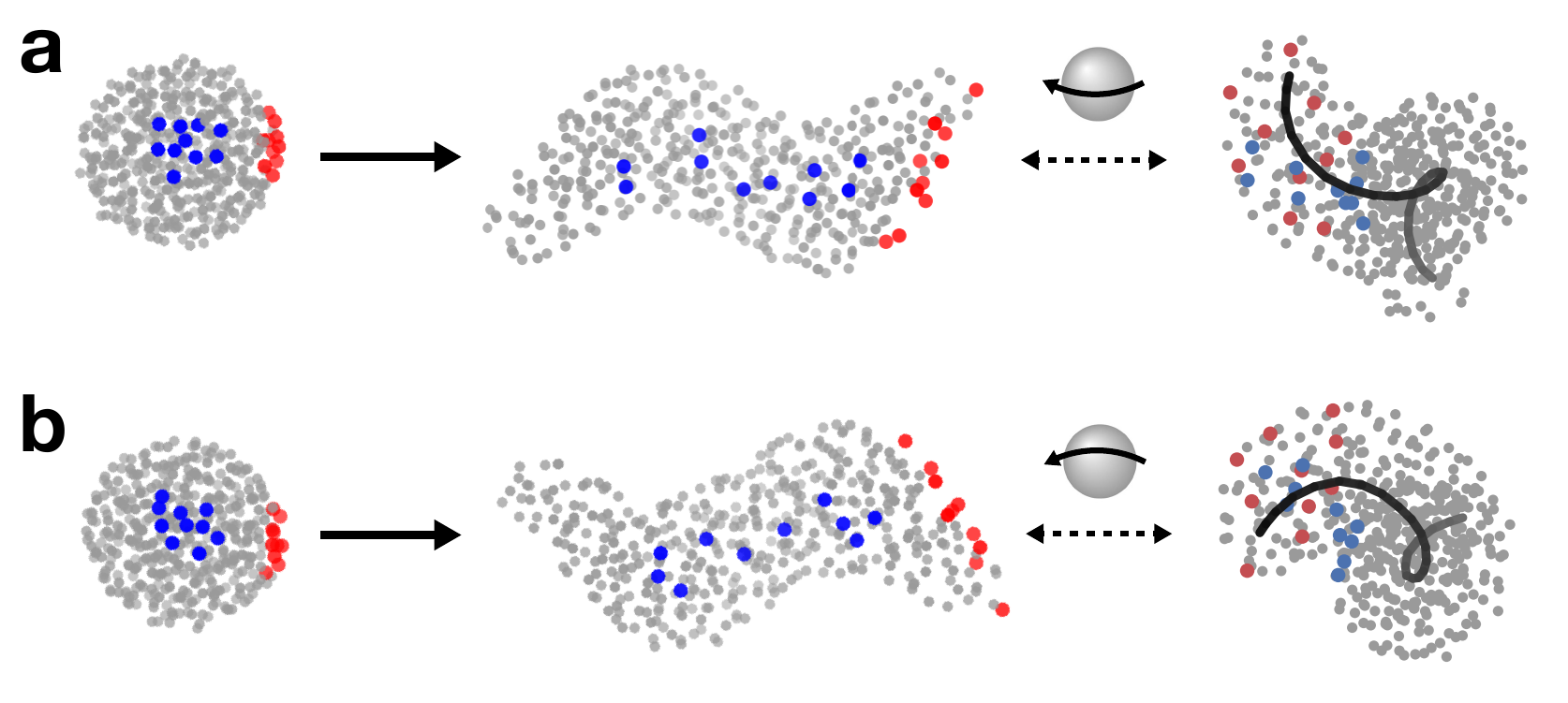}
    \caption{\textbf{Learning helical ellipsoid with two organizers.} (a) The model can learn to form the helical ellipsoid with the correct left-handed chirality from two organizer groups. (b) The same model, when given the initial structure with rotated organizers, fails to reproduce the same chirality. Both the initial and the final shapes are reoriented for comparison against the training orientation in (a). } 
    \label{figure:two_organizers}
\end{figure}
To test whether chirality can be specified without sufficient organizer cues, we train the model to form a left-handed helical ellipsoid using two organizer groups instead of the three used in Figure~\ref{main:fig4} of the main text. As shown in Figure~\hyperref[figure:two_organizers]{S6a}, the model successfully generates the target shape in the training orientation. However, under rotated organizer configurations, it often produces a helical ellipsoid with the opposite handedness---an example is shown in Figure~\hyperref[figure:two_organizers]{S6b}. This suggests that symmetry breaking in this setting is driven by noise-induced irregularities rather than organizer cues. By contrast, the model trained with three organizers reliably produces a left-handed chiral ellipsoid across organizer rotations.

\subsection{Learning shape associated with general tensor-valued weights}
\subsubsection{Implementation}
\label{supp_sec:additional:vector}
\begin{figure}[t!]
    \centering
    \includegraphics[width=0.8\linewidth]{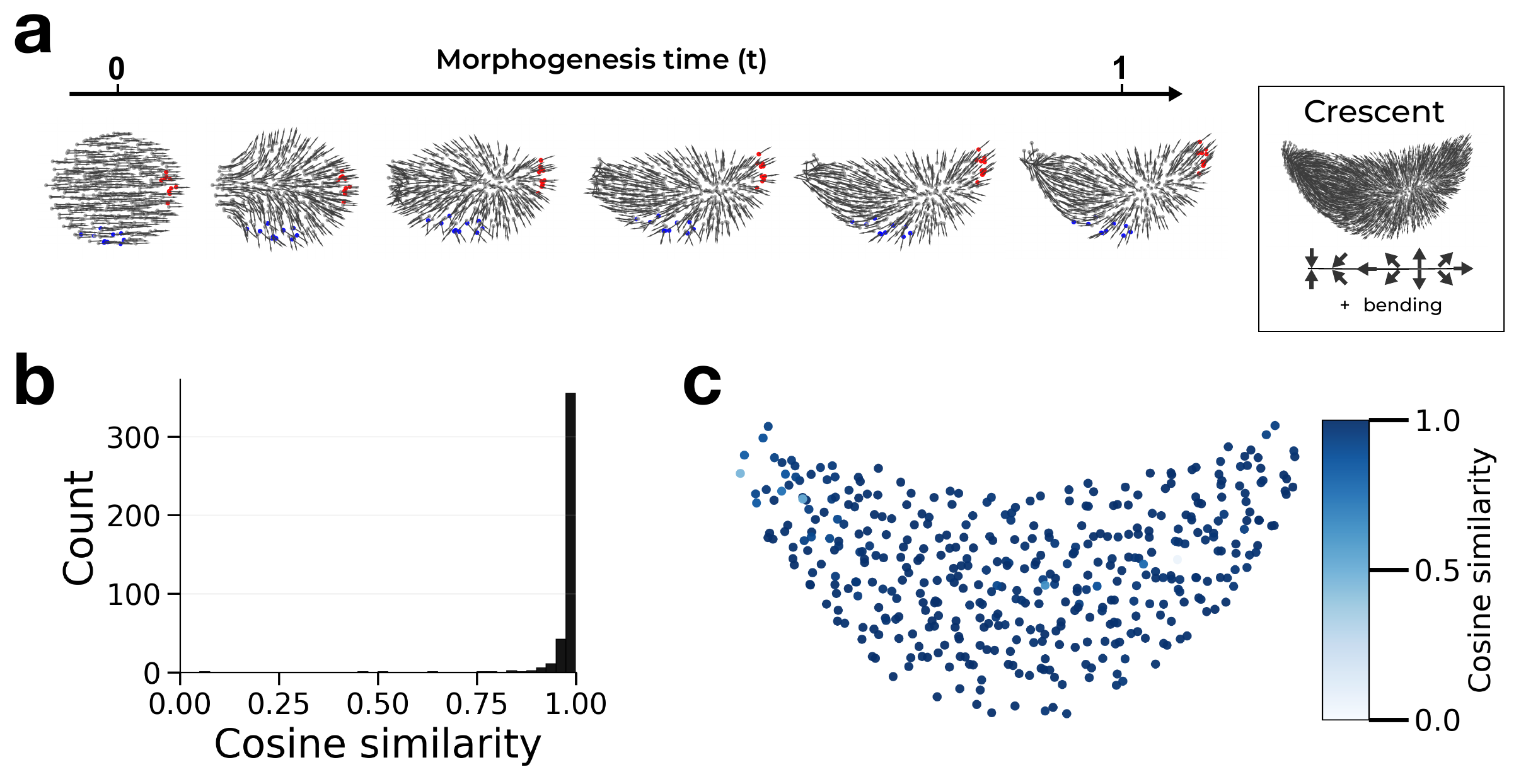}
    \caption{\textbf{Joint shape and director-field learning.} (a) When trained to match the spectra of both the target shape and director vectors, the model forms a crescent morphology while transforming the initially uniformly aligned vectors into the target director arrangement. (b) Histogram of cosine similarity between the learned and target director vectors. (c) Spatial heatmap of cosine similarity over the final morphology. } 
    \label{figure:director_learning}
\end{figure}
We can generalize the weight associated with each point from a scalar to a vector or any higher-order tensor. Since these objects transform under rotation, the spatial (i.e., Cartesian) index of the corresponding Zernike moments should also be transformed away via the spatial rotation matrix \(R\). For example, the \(\ell\)-th order Zernike moments \(\mathbf{z}_\ell\) obtained from vector-valued weights \(\Omega{=}\{(\omega_x, \,\omega_y,\, \omega_z)\}\) have the azimuthal index \(m\) and a spatial index \(i \!\in\! \{x, y, z\}\). Each of them can be rotated with the spectral rotation matrix \(D\) and the spatial rotation matrix \(R\)---which are parametrized by the same unit quaternion---as follows,      
\[
  c'_{n\ell m i} =
    \sum_{m'=-\ell}^{\ell}
    \sum_{j=1}^{3}
    D^\ell_{m m'}(\mathbf{q}) \,
    R_{ij}(\mathbf{q}) \,
    c_{n\ell m' j}
\]
For the higher-order tensor with both upper (contravariant) and lower (covariant) indices, we can use \(R\)---as (1, 1) tensor---and \(R^{-1}\!=\!R^\top\) to transform the respective indices. This formulation enables learning general tensor-valued quantities associated with agents, such as their velocities or stress tensors. 

As a proof of principle, we introduce a vector-valued internal state, polarity \(\mathbf{p}_i\) to each agent \(i\) and evolve it equivariantly (Equation 7 of \citet{satorras2021n}) alongside the positions and gene as,
\begin{equation}
  \dot{\mathbf{p}_i} = \mathbf{f}_i^\mathbf{p}= 
    \phi_\mathbf{p} (\mathbf{g}_i\Vert \mathbf{m}_i)\mathbf{p}_i +  \frac{1}{N-1}\sum_{j\neq i} \phi_\mathbf{p}^\mathbf{x}(\mathbf{m}_{ij}) (\mathbf{x}_i - \mathbf{x}_j)
\end{equation}
and project each spatial component at the final time onto 3D Zernike polynomials to obtain \(\{c^\text{evol}_{n \ell mi }\}\) where the last index is the spatial index \(i \!\in\! \{x, y, z\}\). We train the model with the following combined loss,
\begin{equation}
  \mathcal{L} \! \left(\mathbf{w}\right) = \frac{\lambda}{N_\text{spec}}
    \sum_{n=0}^{n_\text{max}} 
    \sum_{\ell=0}^{\ell_\text{max}}
    \sum_{m=-\ell}^\ell
    \sum_{i=1}^3
      \big\| 
        c^\text{target}_{n \ell m i} \!- \!
        \sum_{m'=-\ell}^\ell \sum_{j=1}^3 \! 
        D^\ell_{mm'}(\mathbf{q}^\star(\mathbf{w}))
        R_{ij}(\mathbf{q}^\star(\mathbf{w}))
        c^\text{evol}_{n \ell m' j}
      \big\|^2 + \mathcal{L}_\text{s.m.}
      \label{eq:director_learning}
\end{equation}
where \(\lambda\) is a weighting coefficient that controls the relative contribution of the director-matching loss.

\subsubsection{Learning crescent shape associated with polarity vectors}
Figure~\hyperref[figure:director_learning]{S7a} visualizes the trajectories of both the cell positions and polarity vectors evolved by the model trained with \eqref{eq:director_learning}. The target is a crescent shape whose director vectors rotate radially from $270^\circ$ on the left to $0^\circ$ on the right along the elongation axis. Starting from initially uniform director vectors on the sphere, the model correctly rotates the directors to match those of the target, as verified by the histogram and cosine-similarity heatmap in Figure~\hyperref[figure:director_learning]{S7b,c}. We used \(\lambda{=0.2}\) and the inner alignment optimization was performed only using the shape-matching part of the combined loss. Otherwise, the training was performed under the same setting as in the agent-based crescent shape-learning experiment, with details provided in Section~\ref{method:agent_based_simulation}.
This confirms that our loss can be used to learn distributions of vector-valued quantities in a rotation-invariant manner by accounting for their transformation under rotation through the spatial rotation matrix \(R\).

\subsection{Ablation of the number of agents and the gene representation}

To assess how the number of agents and the dimensionality of gene expression affect training, we trained separate force models to form the ellipsoid and the bunny, varying either the population size, \(N \in \{351,425,645\}\), or the gene-expression dimension, \(d_{\mathbf{g}} \in \{8,16,32\}\), while keeping all other training settings fixed.

\begin{table}[h!]
\centering
\setlength{\tabcolsep}{4pt}
\captionsetup[subtable]{position=top}
\begin{subtable}[t]{0.46\textwidth}
\centering
\caption{Population size}
\label{table:ablation_N}
\resizebox{\linewidth}{!}{
\begin{tabular}{lccc}
\toprule
\textbf{Target} & $N=351$ & $N=425$ & $N=645$ \\
\midrule
Ellipsoid & 0.37{\footnotesize $\pm$0.02} & 0.41{\footnotesize $\pm$0.02} & 0.40{\footnotesize $\pm$0.02} \\
Bunny     & 0.45{\footnotesize $\pm$0.06} & 0.42{\footnotesize $\pm$0.05} & 0.44{\footnotesize $\pm$0.05} \\
\bottomrule
\end{tabular}
}
\end{subtable}
\hspace{1.5em}
\begin{subtable}[t]{0.42\textwidth}
\centering
\caption{Gene dimension}
\label{table:ablation_dg}
\resizebox{\linewidth}{!}{
\begin{tabular}{lccc}
\toprule
\textbf{Target} & $d_g=8$ & $d_g=16$ & $d_g=32$ \\
\midrule
Ellipsoid & 0.40{\footnotesize $\pm$0.02} & 0.40{\footnotesize $\pm$0.02} & 0.41{\footnotesize $\pm$0.04} \\
Bunny     & 0.58{\footnotesize $\pm$0.06} & 0.52{\footnotesize $\pm$0.04} & 0.40{\footnotesize $\pm$0.02} \\
\bottomrule
\end{tabular}
}
\end{subtable}
\caption{\textbf{Effects of population size and
gene-expression dimensionality.}
(a) Models trained with different population sizes
converge to similar losses.
(b) Reducing the gene-expression dimension has little
effect on the ellipsoid but increases the loss for the bunny. All values are reported in units of \(10^{-2}\).}
\label{table:ablations}
\end{table}

 As shown in Table~\ref{table:ablations}(a), the models converged to similar training losses across the three population sizes, suggesting that the insensitivity of our loss to point-cloud size, demonstrated in Figure~\ref{figure:loss_benchmarking}, extends to DiffeoMorph training. In contrast, reducing gene-expression dimensionality increased the training loss for the bunny but had little effect on the ellipsoid (Table~\ref{table:ablations}(b)). This trend suggests that more complex shapes may require a higher-dimensional gene representation. This interpretation is consistent with Figure~\ref{main:fig5}, where the learned gene expression for the ellipsoid is primarily correlated with two eigenmodes---the second and sixth---whereas the bunny exhibits a more distributed correlation pattern. Thus, the tested gene-expression dimensions may already provide redundant capacity for the ellipsoid, while additional dimensions remain useful for representing the bunny.

\section{Analysis of gene expression evolution from NPA model}
\label{NPA_analysis}
\begin{figure}[t!]
    \centering

    \includegraphics[width=0.92\linewidth]{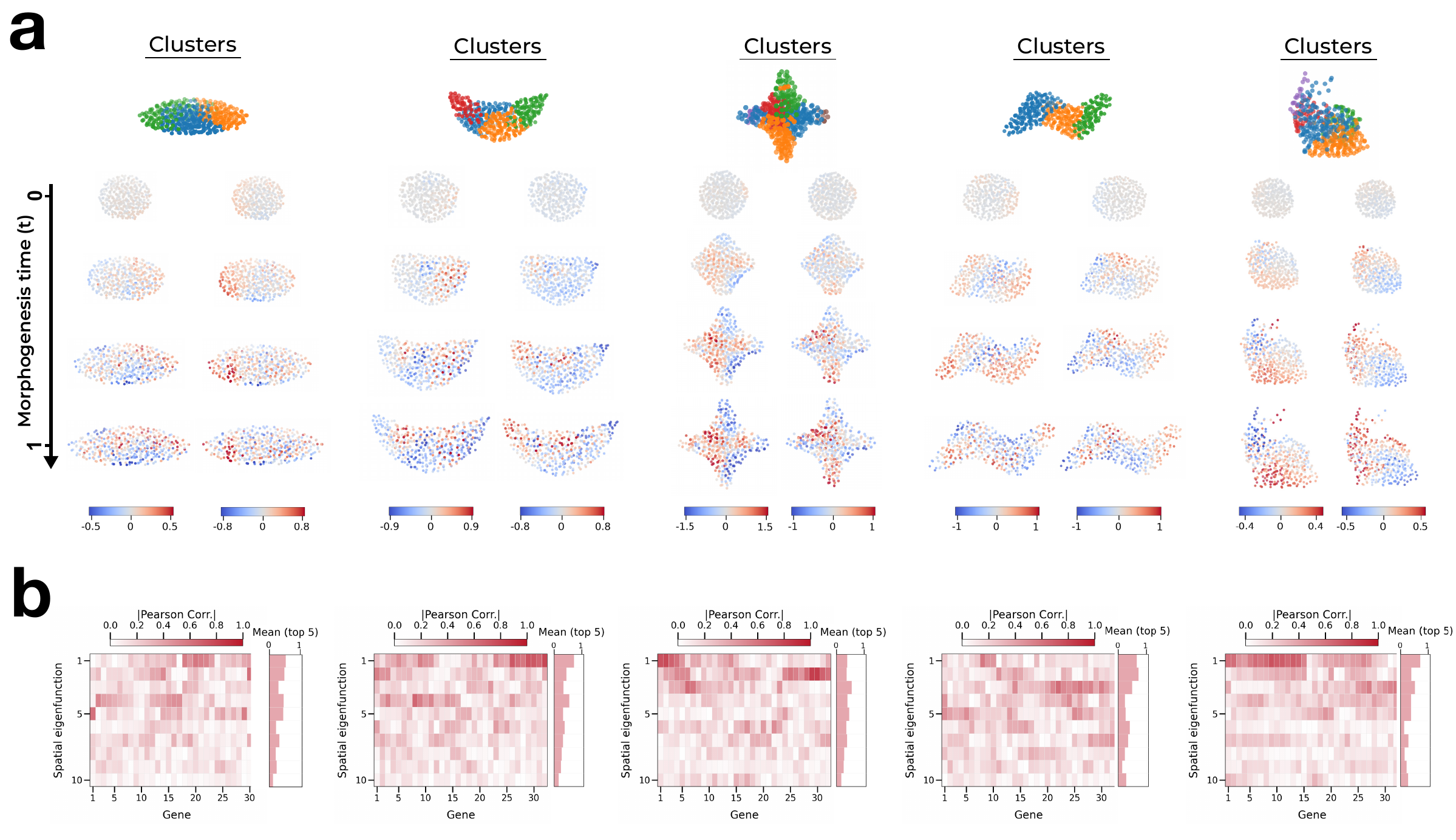}
    
    \caption{\textbf{Gene expression evolution from NPA.} (a) The evolved genes have patches of high and low gene expressions scattered across the shapes, without a clear pattern observed in the equivariant model's evolution. (b) The gene expressions do not show strong correlation with any of the spatial eigenmodes, resulting in a more distributed correlation heatmap compared to the equivariant case.}
    \label{figure:npa_analysis}
\end{figure}

Figure~\ref{figure:npa_analysis} presents the gene expression analysis for the non-equivariant Neural Particle Automata (NPA) model. In contrast to the equivariant model's evolution shown in Figure~\ref{main:fig5} of the main text, the learned gene expression patterns do not organize into coherent spatial domains aligned with the morphology. Although final-time clustering still separates cells into groups, the underlying gene-expression trajectories remain spatially irregular, with high- and low-expression regions scattered across the shape rather than forming smooth compartment-like patterns. The correlation with Laplace--Beltrami eigenmodes, shown as vertical strips next to heatmaps, is also more diffuse, lacking the narrow high-correlation bands observed in the equivariant model. These results suggest that equivariant architecture is essential for discovering a symmetry-aware intrinsic encoding of geometry. Without this inductive bias, the model may instead exploit orientation-specific features, such as noise-induced positional irregularities, rather than learning primarily from organizer cues. This interpretation is consistent with its poor rotational generalization performance shown in Table~\ref{table:model_benchmarking}. 

\begin{wraptable}{r}{0.4\textwidth}
\centering
\resizebox{\linewidth}{!}{
\begin{tabular}{lcc}
\toprule
 & \textbf{NPA} & \textbf{DiffeoMorph} \\
\midrule
Ellipsoid & 0.931 & 0.685 \\
Crescent  & 0.898 & 0.792 \\
Starfish  & 0.985 & 0.813 \\
Helical Ellipsoid     & 0.925 & 0.875 \\
Bunny     & 0.961 & 0.912 \\
\bottomrule
\end{tabular}
}
\caption{\textbf{Entropy of gene–mode correlations.} DiffeoMorph exhibits lower normalized entropy than NPA across all five shapes, indicating more selective alignment with intrinsic spatial modes.}
\label{table:structure_comparison}
\end{wraptable}

Next, we compute normalized entropy using the vertical strips next to the eigenmode-gene correlation heatmap from Figures~\ref{main:fig5} and \ref{figure:npa_analysis}, which summarize the mean of the five largest absolute gene correlations for each mode, to quantify the concentration of gene-eigenmode correlations. Lower entropy means that gene expression aligns selectively with a small number of intrinsic spatial modes whereas higher entropy indicates a more diffuse representation.

As shown in Table~\ref{table:structure_comparison}, DiffeoMorph produces much lower normalized entropy than NPA across all shapes. The differences are especially pronounced for the ellipsoid, crescent, and starfish, which use up to two organizers and thus are missing complete specification of coordinate frames. This association suggests that combining rotation-invariant shape matching with equivariant dynamics encourages the learned gene fields to align with intrinsic Laplace–Beltrami modes, providing the agents with a geometry-adapted internal coordinate system.

\section{Generality of spectral loss beyond agent-based morphogenesis}

\paragraph{Point-cloud completion.}
We trained a PointNet-style network to reconstruct complete chairs from partial scans while independently rotating only the supervision targets. Because target orientation is unpredictable from the input, unaligned Chamfer encourages contracted predictions that average over incompatible poses. DiffeoMorph instead compares intrinsic geometry after spectral alignment. We evaluate predictions using pose-normalized Chamfer$^2$, defined as the sum of the mean squared nearest-neighbor distances in both directions after rotation alignment. As shown in Table~\ref{table:loss_comparison}, DiffeoMorph reduces this error by approximately 51\% relative to unaligned Chamfer and 25\% relative to Chamfer using the same spectral pose estimate. The prediction-to-target radius ratio also improves from 0.548 to 0.995, indicating that DiffeoMorph avoids the severe contraction observed with unaligned supervision.

\paragraph{Superquadric recovery.}
We also tested direct recovery of five intrinsic superquadric parameters---three axis lengths and two shape exponents---from noisy observations with unknown rotation and translation. As shown in Table~\ref{table:fitting_comparison}, across 48 objects, DiffeoMorph reduces dense radial RMSE by 59.0\% relative to jointly optimizing shape and rotation with Chamfer, and by 61.3\% relative to Chamfer with an internally optimized rotation. After averaging matched initializations, DiffeoMorph achieves lower error than both baselines on every object. Retaining radial moments further improves accuracy over the angular-only ablation.

\begin{table}[t]
\centering
\resizebox{\textwidth}{!}{%
\begin{tabular}{lcccc}
\toprule
\textbf{Training loss} & \textbf{Pose-normalized test $\text{Chamfer}^2$} $\downarrow$ & \textbf{95\% CI} & \textbf{Test F-score@0.04} $\uparrow$ & \textbf{Pred./target radius ratio} \\
\midrule
Chamfer without alignment              & 0.03549          & 0.03388--0.03718 & 0.0598          & 0.548 \\
Chamfer after spectral pose alignment  & 0.02325          & 0.02237--0.02418 & 0.1623          & 0.870 \\
DiffeoMorph spectral loss              & \textbf{0.01745} & 0.01649--0.01845 & \textbf{0.1986} & \textbf{0.995} \\
\bottomrule
\end{tabular}%
}
\caption{\textbf{Point-cloud completion with unpredictable target pose.} Results are averaged over three training seeds and 100 held-out chairs, with eight views averaged within each chair. All predictions are evaluated using the same rotation-alignment procedure. The 95\% confidence intervals are obtained from crossed bootstrap resampling of training seeds and test shapes.}
\label{table:loss_comparison}
\end{table}

\begin{table}[t]
\centering
\resizebox{\textwidth}{!}{%
\begin{tabular}{lcccc}
\toprule
\textbf{Fitting objective} & \textbf{Dense radial RMSE} $\downarrow$ & \textbf{95\% CI} & \textbf{Success below 1\%} $\uparrow$ & \textbf{Success below 2\%} $\uparrow$ \\
\midrule
Raw Chamfer without alignment            & 0.14194          & 0.12969--0.15390 & 0.0\%           & 0.0\% \\
Joint shape-and-rotation Chamfer         & 0.01372          & 0.01199--0.01562 & 36.8\%          & 81.2\% \\
Rotation-aligned Chamfer                 & 0.01455          & 0.01276--0.01654 & 32.6\%          & 76.4\% \\
\midrule
DiffeoMorph, angular-only ablation       & 0.00774          & 0.00551--0.01092 & 81.2\%          & 95.8\% \\
DiffeoMorph, angular and radial moments  & \textbf{0.00563} & 0.00466--0.00663 & \textbf{86.8\%} & \textbf{99.3\%} \\
\bottomrule
\end{tabular}%
}
\caption{\textbf{Recovery of intrinsic superquadric parameters.} Relative radial RMSE is evaluated along 2,048 canonical directions for 48 objects. Errors are averaged over three matched initializations per object, and 95\% confidence intervals are obtained by bootstrap resampling objects. Success rates report the fraction of fits below the stated relative-error thresholds, averaged within objects and then across objects.}
\label{table:fitting_comparison}
\end{table}

The advantage in both settings comes from separating intrinsic geometry from nuisance pose. In point-cloud completion, unpredictable target rotations give raw Chamfer incompatible training signals, encouraging contracted, pose-averaged predictions; spectral alignment removes this conflict. In superquadric fitting, Chamfer's nearest-neighbor assignments can switch abruptly as pose and shape change, whereas the global Zernike representation and its Wigner--D transformation vary smoothly. The spectral objective thus offers a smoother optimization signal while retaining radial size information through the nonconstant \(\ell{=}0\) coefficients.

\section{Experimental details}
\label{supp_sec:experimental_details}

\subsection{Computing resources}
All experiments were conducted on two NVIDIA RTX 6000 Ada Generation GPUs, each with 48GB of memory. 

\subsection{Generating shapes}
We prepared an approximately uniformly sampled ball \(\mathbf{X}\) of radius 1 with \(N{=}2000\) cells by sampling random directions and volume-corrected radii. 

\paragraph{Ellipsoid.}
We obtain an ellipsoid by uniformly scaling a ball along the \(x\)-axis by a factor of~3.

\paragraph{Crescent.}
Starting from the ellipsoid, we introduce a bending transformation with angle \(\theta_i {=} x_i / R\), updating coordinates as \(x_i^{\text{new}} {=} R \sin(\theta_i)\), \(y_i^{\text{new}} {=} y_i\), and \(z_i^{\text{new}} {=} R(1 {-} \cos(\theta_i)) + z_i\) with \(R {=} 3\), which produces a crescent shape.

\paragraph{Crescent with polarity vectors.}
Let \(\mathbf{e}_x{=}(1,0,0)^\top\) and \(\mathbf{e}_\perp(\mathbf{x}){=}(0,y,z)^\top/\sqrt{y^2+z^2}\), where \(\mathbf{x}{=}(x,y,z)^\top\), denote the axial direction along the \(x\)-axis and the radial direction in the \((y,z)\)-plane, respectively. We define the director angle as \(\theta(s){=}135^\circ(1-s)\), so that the director rotates from \(270^\circ\) to \(0^\circ\) as the normalized axial coordinate \(s\) varies from \(-1\) to \(1\), corresponding to \(x{\in}[-1.25,1.25]\) for an ellipsoid of length \(2.5\). The pre-bending director vector \(\tilde{\mathbf{p}}(\mathbf{x}_i)\) of cell \(i\), illustrated in the target-shape box of Figure~\hyperref[figure:director_learning]{S7a}, is obtained as
\[
\tilde{\mathbf{p}}(\mathbf{x}_i)
=
\cos\theta(s_i)\mathbf{e}_x
+
\sin\theta(s_i)\mathbf{e}_\perp(\mathbf{x}_i).
\]

Under the bending transformation described above, the pre-bending director vectors are transported by the Jacobian of the bending map,
\[
J_B(\mathbf{x}){=}
\begin{pmatrix}
\cos(x/R) & 0 & 0\\
0 & 1 & 0\\
\sin(x/R) & 0 & 1
\end{pmatrix}.
\]
The director vector on the crescent is then obtained by \(\mathbf{p}(\mathbf{x}_i){=}J_B(\mathbf{x}_i)\tilde{\mathbf{p}}(\mathbf{x}_i)\) and normalizing it to unit norm.

\paragraph{Starfish.}
We generate a starfish shape by deforming the ball in cylindrical coordinates, where \(\theta_i {=} \operatorname{atan2}(y_i, x_i)\) and \(r_i {=} \sqrt{x_i^2 + y_i^2}\). The boundary is defined by an angular-dependent radius \(R(\theta_i) {=} R_0 \left(1 + A[\max(0, \cos(4\theta_i))]^s\right)\), and we enforce \(r_i \leq R(\theta_i)\) together with a thickness constraint \(|z_i| \leq h\left(1 - (r_i / R(\theta_i))^q\right)^{1/q}\). The resulting shape is scaled as \(p_i^{\mathrm{new}} {=} \lambda (x_i, y_i, z_i)\), with parameters \(R_0 {=} 1.0\), \(A {=} 0.8\), \(s {=} 1.0\), \(h {=} 0.7\), \(q {=} 1.5\), and \(\lambda {=} 1.2\), yielding a four-legged starfish.

\paragraph{Helical ellipsoid.}
We construct a helical variant of the ellipsoid by defining a normalized axial coordinate \(t_i {=} (x_i - \operatorname{mean}(x)) / (x_{\max} - x_{\min})\) and a corresponding phase \(\theta_i {=} -2\pi T t_i\). The deformation is applied by keeping \(x_i\) unchanged while shifting the transverse coordinates as \(y_i^{\mathrm{new}} {=} y_i + r \cos(\theta_i)\) and \(z_i^{\mathrm{new}} {=} z_i + r \sin(\theta_i)\), with \(T {=} 1.5\) and \(r {=} 0.5\). The minus sign of the phase sets the left-handed chirality.

\paragraph{Stanford Bunny.}
We downloaded surface-only and volumetric point clouds of the Stanford Bunny from the \href{http://graphics.stanford.edu/data/3Dscanrep/}{Stanford 3D Scanning Repository} and the \href{https://github.com/dcoeurjo/VolGallery}{VolGallery repository}, respectively. The volumetric point cloud was subsampled to 10,000 points and rescaled so that \(r_\text{max}{=}3.5\).

\subsection{Implementation details of spectral projection and rotation}

\paragraph{Projection.}
We precomputed the radial polynomials \(R_{n\ell}\) up to \(n_{\text{max}}{=}20\) and \(\ell_{\text{max}}{=}10\) to avoid repeatedly evaluating terms involving the Jacobi and associated Legendre polynomials. We used the \texttt{e3nn} library~\citep{geiger2022e3nn} to compute the real spherical harmonics \(Y_{\ell m}\), which were assembled into a matrix \(\mathbf{Y}\). We used \texttt{vmap} to vectorize the projection over \(n\), \(\ell\), and \(m\) in Equation~\eqref{eq2}. Since tensors passed to \texttt{vmap} must have identical shapes, for each \(\ell\)-row of \(\mathbf{Y}\), we placed the numerical values of the real spherical harmonics at the center of \(2\ell_{\text{max}}{+}1\) entries, aligning the \(m{=}0\) component with the central column and zero-padding the remaining entries. Thus, \(\mathbf{Y}\) has size \(10{\times}21\), after excluding the scalar components at \(\ell{=}0\). For the radial polynomials, the valid \((n,\ell)\) pairs satisfy \(n{-}\ell\) even and \(n{>}\ell\). We chose \(n_{\text{max}}{=}20\) and, for each \(\ell\), placed the nonzero \(n\)-th radial polynomial values starting from the first column. The resulting \(R_{n\ell}\) array has size \(20{\times}10\).

\paragraph{Spectral Rotation.}
We precomputed the binomial terms and factorial-based prefactors in the Wigner--D matrix expression in Equation~\eqref{supp:eq5} for all \((\ell,m,m',\rho)\) configurations up to \(\ell_{\text{max}}{=}10\), and used these values as lookup tables when evaluating the Wigner--D matrix. We vectorized the computation over \((m,m',\rho)\) at each \(\ell\), as well as over \(\ell\). Since tensors returned by \texttt{vmap} must have identical shapes, for each \(\ell\), we placed the numerical values of the Wigner--D matrix at the center of a \((2\ell_{\text{max}}{+}1){\times}(2\ell_{\text{max}}{+}1)\) array and zero-padded the remaining entries.

\subsubsection{Loss testing}

\paragraph{Spectral alignment.}
We projected the Stanford bunny in Figure~\hyperref[main:fig2]{2a} and the weighted crescent in Figure~\hyperref[figure:weighted_shape]{S5a} onto 3D Zernike polynomials with angular order up to \(\ell_\text{max}{=}8\), and initialized a unit quaternion in a random direction. We then performed gradient ascent with learning rate \(\eta{=}\texttt{5e-3}\) and terminated the optimization when the change in spectral overlap \(\mathcal{M}\) fell below \(\texttt{1e-8}\).

\paragraph{Direct shape learning.}
\label{direct_shape_learning_detail}
We prepared an approximately uniformly sampled ball \(\mathbf{X}\) of radius 1 by sampling random directions and volume-corrected radii. All weights \(\boldsymbol{\omega}\) were initialized to 1. At each training step, we normalized the updated shape by \(r_\text{max}\) of the corresponding target structure: 3 for the crescent in Figure~\hyperref[main:fig2]{2b} and 3.5 for the weighted Stanford bunny in Figure~\hyperref[figure:weighted_shape]{S5b}. We then projected the normalized point cloud onto 3D Zernike polynomials with angular order up to \(\ell_\text{max}{=}10\). The unit quaternion for the inner alignment optimization was randomly initialized and optimized with learning rate \(\eta_\text{inner}{=}\texttt{1e-1}\) until the change in the inner objective fell below \(\delta_\text{inner}{=}\texttt{1e-8}\). At each outer training step, we warm-started the inner optimization from the unit quaternion obtained in the previous iteration. For the outer optimization, we learned the point coordinates \(\mathbf{X}\) and, in the bunny experiment, the weights \(\boldsymbol{\omega}\), using learning rates \(\eta_{\mathbf{X}}{=}\eta_{\boldsymbol{\omega}}{=}\texttt{5e-2}\). We terminated training when the shape-matching loss \(\mathcal{L}_\text{s.m.}\) fell below \(\delta_\text{outer}{=}\texttt{5e-5}\).

\subsection{Agent-based simulation}
\label{method:agent_based_simulation}

\subsubsection{Preparation of initial structure}
In DiffeoMorph, we learn a distributed morphogenesis control protocol based on the evolving states of individual agents. To make the learning problem biologically realistic and algorithmically nontrivial, we initialize the positions and gene expression profiles in a way that provides only minimal spatial cues so that effective morphogenesis requires inter-agent communication. 

\paragraph{Positions.}
\label{method:initialization:positions}
The sampling scheme used for direct shape learning above can introduce substantial irregularities throughout the volume. To prepare a more uniform sphere-shaped point cloud \(\mathbf{X}^0\), we sampled the shell and core of the sphere separately. The shell was generated using a Fibonacci lattice \citep{gonzalez2010measurement} with 250 agents, providing a quasi-uniform discretization of \(S^2\) with mean pairwise distance \(\langle r_{ij}\rangle{=}0.21\). To obtain a core with inter-agent spacing comparable to \(\langle r_{ij}\rangle\), we sampled points inside a sphere of radius \(1{-}\langle r_{ij}\rangle\) using Poisson-disk sampling \citep{bridson2007fast}, with the minimum separation distance set to \(\langle r_{ij}\rangle\). This resulted in 175 core agents. Although no algorithm can produce a perfectly uniform grid in the unit ball, combining the shell and core samples yields a point cloud that approximates a uniform distribution over the unit ball with \(N_\text{agent}{=}425\). The point cloud was then elongated by \(10\%\) along a selected axis; the motivation for this elongation and the choice of axis are described in the next section. In addition, we perturbed the point cloud at every training step by adding zero-mean Gaussian noise, denoted by \(\epsilon\) in Figure~\ref{main:fig4}. The prescribed noise magnitude was 0.05 for the ellipsoid and crescent experiments, and 0.03 for the starfish, helical ellipsoid, and bunny experiments. This perturbation encourages the force model to learn a generalizable morphogenesis protocol.

\paragraph{Gene expression profiles.} 
\label{method:initialization:gene}
Taking inspiration again from developmental biology, we interpret the elongation as originating from a preceding, localized morphogen expression at a pole, which establishes the elongation axis and thereby sets the body axis \citep{wolpert2015principles}. To encode this spatial cue on the point cloud, we randomly sample a unit vector as body axis, project all cells onto this vector, and identify the cell with the maximal projection---this marks the pole. We then select its \( N_\text{org}{-}1 \) nearest neighbors on the shell to define \( N_\text{org}\) ``organizer cells" expressing the morphogens. These cells are assigned the one-hot gene expression \( \mathbf{g}_i {=} [1,0,\dots,0] \), while all remaining cells are assigned \( \mathbf{g}_i {=} [0,\dots,0, 1]\). The point cloud is then elongated along the sampled direction. We set the dimensionality of the gene expression vector \(d_\mathbf{g}{=}32\). When introducing additional groups of organizer cells, we sample new unit vectors that are \( 90^\circ \) apart from the previously chosen pole axis (with the third axis determined by the right-hand rule), and again select \( N_\text{org}\) organizer cells around each new pole. The second group receives \( \mathbf{g}_i {=} [0, 1, 0, \dots,0] \), and the third receives \( \mathbf{g}_i {=} [0, 0, 1, 0, \dots, 0] \). These assignments encode morphogen signals that define left-right and front-back symmetry axes. In contrast to the primary body axis, no perturbation is applied along these additional directions.

\subsubsection{Details of model architecture and training}
\label{experimental_detail:architecture_and_training}
All MLPs in the force model had three hidden layers with 32 units. We used the \texttt{SiLU} activation function \citep{ramachandran2017searching} to obtain a smooth, numerically stable vector field. For numerical integration, we used the \texttt{Euler} solver from \texttt{Diffrax} \citep{kidger2021on} with timestep \(\texttt{dt=0.01}\). For training, we used \(\eta_\text{inner}{=}\texttt{1e-2}\) and \(\delta_\text{inner}{=}\texttt{1e-7}\) for spectral alignment, and \(\eta_\text{outer}{=}\texttt{5e-3}\) for force model training. We warm-started the inner alignment optimization as in the direct shape learning setup. Training was early-stopped if the loss did not improve for 500 steps for the starfish and 250 for the rest. Each evolved point cloud was normalized by a target-specific radius \(r_\text{max}\), initialized at \(1.5\) and increased by \(0.05\) whenever the evolved structure's radius came within \(0.4\) of the current value. This schedule was capped at \(r_\text{max}{=}3\) for the ellipsoid, crescent, and helical ellipsoid, \(2.1\) for the starfish, and \(3.5\) for the bunny. We used \(N_\text{org}{=}10\) for all shapes except the bunny, where \(N_\text{org}{=}20\). During evolution, we added Brownian noise with magnitude \(\sigma_{\mathbf{x}}{=}0.002\) to the positions and set \(\sigma_\mathbf{g}{=}0\), making the gene expression dynamics deterministic.

\subsubsection{Generalization experiments}
For noise generalization, we kept the organizer placements fixed as in training and perturbed the initial point cloud with noise magnitudes 0.05, 0.1, and 0.2. For rotational generalization, at each noise magnitude, we sampled rotation angles and applied the corresponding rotations to the organizers. The MSE loss for the generalization under organizer rotation shown in Table~\ref{table:model_benchmarking} was evaluated by performing a fresh spectral alignment optimization for each test sample. For the ellipsoid and crescent experiments, we warm-started this optimization using the optimal unit quaternion \(\mathbf{q}^\star\) obtained during training. In the bunny experiment, however, we observed that initializing with \(\mathbf{q}^\star\) sometimes resulted in a high test loss---even when the evolved shape visually reproduces the bunny---suggesting that the alignment optimization, starting from the training orientation, became trapped in a local maximum due to the high-frequency modes of the bunny morphology. To address this, we reorient \( \mathbf{q}^\star \) by first computing the quaternion \(\mathbf{q}_{\text{1}\rightarrow \text{0}}\) that reverses the organizer shift---from the shifted direction \(\mathbf{u}_\text{1}\) back to the original direction \(\mathbf{u}_\text{0}\), which defines the regions of the red organizer cells in Figure~\hyperref[main:fig4]{3a}---as follows:
\[
  \mathbf{q}_{\text{1}\rightarrow \text{0}} =
    \left( 
      \cos \frac{\theta}{2}, \mathbf{v}\sin \frac{\theta}{2}
    \right)  
    \quad \text{where} \quad 
    \theta=\arccos(\mathbf{u}_\text{1} \cdot \mathbf{u}_\text{0}) 
    \ \ \text{and} \ \
    \mathbf{v} = \frac{\mathbf{u}_\text{1} \!\times \!\mathbf{u}_\text{0} }{\Vert\mathbf{u}_\text{1} \!\times \!\mathbf{u}_\text{0} \Vert}
\]
and then updated the unit quaternion used for the loss evaluation to \(\mathbf{q}_\text{new} = \mathbf{q}^\star \otimes \mathbf{q}_{\text{1}\rightarrow \text{0}}\) where \( \otimes \) denotes quaternion multiplication.

\subsection{Learning oscillatory behavior}
As illustrated in Figure~\hyperref[main:fig4]{3b}, we performed shape matching against an ellipsoid of length~2 at \(t{=}0.5,\,1,\,1.5\) and length~3 at \(t{=}0.75,\,1.25\), thereby covering two periods of alternation between the short and long ellipsoidal shapes. At \(t{=}0.5\), we used the standard outer-loss computation to learn the optimal unit quaternion. For subsequent matching times, we evaluated the losses using this learned quaternion while re-optimizing only the angle. This choice is motivated by the fact that once the model learns to elongate along a given direction during the first shape-matching step, the same orientation can be retained for the remaining targets. To stabilize the resulting oscillatory dynamics, we additionally introduced a velocity consistency loss,
\begin{equation}
\mathcal{L}_\text{v.c.} = \frac{1}{N_\text{agent}}\sum_{t\in\mathcal{I}}\Big(\big\Vert \dot{\mathbf{X}}^t-\dot{\mathbf{X}}^{t+\tau}\big\Vert^2 + \big\Vert \dot{\mathbf{G}}^t-\dot{\mathbf{G}}^{t+\tau}\big\Vert^2\Big).
\label{velocity_consistency_loss}
\end{equation}

We used \(\mathcal{I}{=}\{0.625,\,0.875\}\), corresponding to the midpoints of the transitions from the short to long form and from the long to short form, respectively, and set \(\tau{=}0.5\). We then differentiated the sum of the shape-matching and the velocity consistency losses across the selected time points. The training setup was otherwise identical to that used for crescent learning, as described in Section~\ref{experimental_detail:architecture_and_training}, except that we reduced the noise magnitude of the initial position perturbation to \(0.02\) and set \(\sigma_{\mathbf{x}}{=}0\). For reintegration, we increased the initial noise magnitude back to \(0.05\).

\subsection{Learning stationary shape}
To form a stationary shape at \(t{=}t_f{=}1\), we include the velocity penalty term,
\begin{equation}
    \mathcal{L}_\text{v.p.}=\frac{1}{N_\text{agent}}\Big(\big\Vert \dot{\mathbf{X}}^{t_f}\big\Vert^2 + \big\Vert \dot{\mathbf{G}}^{t_f}\big\Vert^2\Big)
\end{equation}
alongside the shape-matching loss for the outer optimization. The training setup was otherwise identical to that used for bunny learning in Section~\ref{experimental_detail:architecture_and_training}.

\subsection{Gene analysis}
Using the pairwise distances \(d_{ij}{=}\Vert \mathbf{x}_i{-}\mathbf{x}_j\Vert\), we constructed a \(k\)-nearest-neighbor graph with \(k{=}12\). We then computed the sparse affinity matrix \(\mathbf{W}{\in}\mathbb{R}^{N_{\text{agent}}\times N_{\text{agent}}}\) as
\[
[\mathbf{W}]_{ij} =
\begin{cases}
\exp\!\left(-\dfrac{d_{ij}^2}{\sigma^2}\right) & j \in \mathcal{N}_k(i)\\[0.6em]
0 & \text{otherwise}
\end{cases}
\]
where \(\mathcal{N}_k(i)\) denotes the set of \(k\)-nearest neighbors of agent \(i\), and \(\sigma\) was set to the median pairwise distance over all edges in the \(k\)NN graph.

We symmetrized the affinity matrix by setting
\[
[\mathbf{W}]_{ij}{\leftarrow}\max\!\big([\mathbf{W}]_{ij},[\mathbf{W}]_{ji}\big),
\]
and computed the diagonal degree matrix
\[
[\mathbf{D}]_{ii} = \sum_{j=1}^{N_{\text{agent}}}[\mathbf{W}]_{ij}.
\]
The symmetric normalized graph Laplacian was then defined as
\[
\mathbf{L}{=}\mathbf{I}-\mathbf{D}^{-1/2}\mathbf{W}\mathbf{D}^{-1/2},
\]
with a small regularization parameter \(\epsilon{=}\texttt{1e-8}\) added to the degrees for numerical stability. This Laplacian provides a discrete analogue of the Laplace--Beltrami operator on the shape manifold.

The spatial eigenmodes shown in Figures~\hyperref[main:fig5]{4b} and \hyperref[figure:npa_analysis]{S8b} are the eigenvectors \(\mathbf{u}_k{\in}\mathbb{R}^{N_{\text{agent}}}\) of \(\mathbf{L}\). For each eigenmode \(\mathbf{u}_k\), we computed its Pearson correlation with each raw gene-expression vector \(\mathbf{g}_k{\in}\mathbb{R}^{N_{\text{agent}}}\), defined as the \(k\)-th column of \(\mathbf{G}^{t_f}{\in}\mathbb{R}^{N_{\text{agent}}\times d_{\mathbf{g}}}\), without further processing.

\subsection{Point-cloud completion and superquadric recovery}
\paragraph{Point-cloud completion.}
We used ModelNet40 chair meshes, with 800 shapes for training, 89 for validation, and 100 held-out test shapes. Partial scans were generated by retaining the camera-facing half of an area-sampled mesh surface. Each example contained 512 partial-input points and 512 independently sampled complete-target points, with four, two, and eight views per shape for training, validation, and testing, respectively. The partial input remained in its dataset frame, while the complete target received an independent Haar-uniform rotation in \(\mathrm{SO}(3)\). We used a PointNet-style encoder with shared MLP layers \(3\to64\to128\to256\) and max pooling, followed by a decoder \(256\to256\to512\to1536\), producing 512 three-dimensional points. Each objective was trained for 3,000 AdamW updates using three matched seeds and the final checkpoint. All predictions were evaluated using the same hybrid rotation search, selecting the alignment that minimized centered symmetric squared Chamfer distance. Metrics were averaged over views within each test shape, and 95\% confidence intervals were computed using 10,000 crossed bootstrap resamples of training seeds and test shapes.

\paragraph{Superquadric recovery.}
We generated 48 independent triaxial convex superquadrics with axis lengths \(a_x\in[0.65,0.85]\), \(a_y\in[0.45,0.62]\), and \(a_z\in[0.28,0.43]\), and exponents \(\epsilon_1,\epsilon_2\in[0.45,1.45]\). Each observation contained 256 points, subjected to a Haar-uniform rotation, random translation with extent 0.35, and Gaussian coordinate noise with standard deviation 0.002. After centering, we fitted a differentiable 192-point surface using three matched parameter initializations per object and 400 Adam updates with learning rate 0.035. The joint-Chamfer baseline optimized shape and rotation together, whereas the aligned-Chamfer baseline optimized rotation internally using the Chamfer objective. For DiffeoMorph, rotation was fitted in Zernike space and held fixed during the outer parameter update. Moment normalizers were estimated once from the unlabeled confirmation cohort and held fixed during fitting. We evaluated relative radial RMSE along 2,048 independent canonical directions, averaged errors over initializations within each object, and computed 95\% confidence intervals using 10,000 object-level bootstrap resamples.

\section{Limitations}
\paragraph{Scalability of the force model.}
The current implementation uses dense communication, giving \(O(N^2)\) message-passing cost per simulation step. This is sufficient for the agent counts studied here, but scaling to much larger populations will require sparse neighborhoods, radius cutoffs, or fast approximate interaction schemes. Importantly, this limitation concerns the force-model implementation rather than the spectral shape-matching objective, which is computed after projection to the Zernike moments.

\paragraph{Existence of local optima in spectral alignment.} The loss landscape of spectral overlap \(\mathcal{M}\) may contain suboptimal local maxima, particularly when comparing shapes with high-frequency modes and using a sufficiently high \(\ell_\text{max}\) to capture these variations. This issue is partially mitigated in DiffeoMorph setup where we start with a highly symmetric 3D shape and hence effectively compare only low-frequency modes in the early stages. Another practical strategy is to use low-order Zernike moments to provide a smoother loss landscape for the inner alignment optimization while evaluating the outer loss on the full set of moments. In this case, the envelope-theorem simplification no longer applies directly, because the inner objective differs from the \(\mathbf{q}\)-dependent part of the outer loss. Consequently, \(\mathbf{q}^\star\) cannot be detached and the Jacobian term must be accounted for during the outer optimization. This can be done efficiently with the proposed implicit differentiation.

\section{Broader impacts}
DiffeoMorph is a foundational framework for learning distributed control rules for morphogenesis-like shape formation. Its potential positive impacts include providing computational tools for studying developmental principles, designing programmable self-assembly systems, and improving distributed control strategies for swarm robotics and programmable matter. At the same time, methods for controlling collective biological or robotic systems may carry risks if applied prematurely or without appropriate safeguards. In biological contexts, extensions toward real morphogenetic or tissue-engineering systems would require careful experimental validation, biosafety assessment, and domain-specific oversight.

\section{Mathematical expressions of benchmark distance metrics}
\label{supp_sec:expressions_of_metrics}

We summarize here the mathematical expressions of the distance metrics compared in Section~\ref{supp_sec:loss_benchmarking} for the readers' convenience. Let \(N\) and \(M\) denote the number of points in coordinate matrices \(\mathbf{X}\) and \(\mathbf{Y}\) respectively, and \( \vert \cdot \vert \) denotes cardinality of a set.

\begin{equation*}
\begin{aligned}
  \textbf{Chamfer :} \quad & 
    \frac{1}{N} \sum_{i=1}^N \min_{j} \lVert \mathbf{x}_i-\mathbf{y}_j \rVert^2 + \frac{1}{M} \sum_{j=1}^M \min_{i} \lVert \mathbf{y}_j-\mathbf{x}_i \rVert^2  \\
  \textbf{Earth Mover's :} \quad & 
    \underset{\boldsymbol{\pi} \in \boldsymbol{\Pi}(\mathbf{a}, \mathbf{b})}{\operatorname{min}} \sum_{i=1}^N \sum_{j=1}^M \boldsymbol{\pi}_{ij} \lVert \mathbf{x}_i - \mathbf{y}_j \rVert  \\
  \textbf{Pairwise :} \quad & 
    \frac{1}{N^2} \sum_{i,j=1}^N \big( \lVert \mathbf{x}_i - \mathbf{x}_j \rVert - \lVert \mathbf{y}_i - \mathbf{y}_j \rVert \big)^2 \\
  \textbf{Gromov--Wasserstein :} \quad & 
    \underset{\boldsymbol{\pi} \in \boldsymbol{\Pi}(\mathbf{a}, \mathbf{b})}{\operatorname{min}} \sum_{i,i'=1}^N \sum_{j,j'=1}^M \boldsymbol{\pi}_{ij} \boldsymbol{\pi}_{i'j'}\big( \lVert \mathbf{x}_i - \mathbf{x}_{i'}\rVert - \lVert \mathbf{y}_j - \mathbf{y}_{j'} \rVert \big)^2 \\
  \textbf{Power Spectrum :} \quad & 
    \frac{1}{n_\text{max} \, \ell_\text{max}} \sum_n^{n_\text{max}} \sum_\ell^{\ell_\text{max}}  \left(   \mathbf{A}_{n\ell}^{\mathbf{X}}  -  \mathbf{A}_{n\ell}^{\mathbf{Y}}\right)^2\\
  \textbf{Bispectrum :} \quad & 
    \frac{1}{n_\text{max} \, \vert \mathcal{I}_\mathcal{B}\vert} \sum_n^{n_\text{max}} \sum_{(\ell_1,\ell_2,\ell_3)\in \mathcal{I}_\mathcal{B}}  \left( \mathbf{B}_{n \ell_1 \ell_2 \ell_3}^{\mathbf{X}} - \mathbf{B}_{n \ell_1 \ell_2 \ell_3}^{\mathbf{Y}}  \right)^2 \\
  \textbf{Trispectrum :} \quad & 
    \frac{1}{n_\text{max} \, \vert \mathcal{I}_\mathcal{T}\vert} \sum_n^{n_\text{max}} \sum_{(\ell_1,\ell_2,\ell_3,\ell_4)\in \mathcal{I}_\mathcal{T}} \left( \mathbf{T}_{n \ell_1 \ell_2 \ell_3 \ell_4}^{\mathbf{X}} - \mathbf{T}_{n \ell_1 \ell_2 \ell_3 \ell_4}^{\mathbf{Y}}  \right)^2  \\
\end{aligned}
\end{equation*}

In the Earth Mover's and Gromov--Wasserstein distances, \(\boldsymbol{\Pi}(\mathbf{a},\mathbf{b})\) denotes the set of all discrete transport plans (i.e., probabilistic coupling matrices) between two discrete mass vectors \(\mathbf{a}=\left(\frac{1}{N},\dots \frac{1}{N}\right)^{\!\top}\) and \(\mathbf{b}=\left(\frac{1}{M},\dots \frac{1}{M}\right)^{\!\top}\). \(\mathcal{I}_\mathcal{B}\) of Bispectrum refers to the set of all triplets \((\ell_1, \ell_2, \ell_3)\) of angular degree combinations that satisfy the selection rules:
\begin{equation*}
\begin{aligned}
  \text{Angular momentum coupling :} &\quad |\ell_1 - \ell_2| \le \ell_3 \le \ell_1 + \ell_2 \\
  \text{Parity constraint :} &\quad \ell_1 + \ell_2 + \ell_3 = \text{Even Integer}
\end{aligned}
\end{equation*}

Similarly, \(\mathcal{I}_\mathcal{T}\) is the set of all quartets \((\ell_1, \ell_2, \ell_3, \ell_4)\), together with \(\ell'\!\in\!\mathcal{L}(\ell_1, \ell_2, \ell_3, \ell_4)\), that satisfy the selection rules:
\begin{equation*}
\begin{aligned}
  \text{Angular momentum coupling :} &\quad |\ell_1 - \ell_2| \le \ell' \le \ell_1 + \ell_2  \quad \text{and} \quad |\ell_3 - \ell_4| \le \ell' \le \ell_3 + \ell_4 \\
  \text{Parity constraint :} &\quad \ell_1 + \ell_2 + \ell' = \ell_3 + \ell_4 + \ell' = \text{Even Integer}
\end{aligned}
\end{equation*}

The two-, three-, and four-point invariants \( \mathbf{A}^{(\cdot)}_{n\ell}\), \(\mathbf{B}_{n \ell_1 \ell_2 \ell_3}^{(\cdot)}\), and \(\mathbf{T}_{n \ell_1 \ell_2 \ell_3 \ell_4}^{(\cdot)}\) in Power spectrum, Bispectrum, and Trispectrum are:
\begin{equation*}
\begin{aligned}
  &\mathbf{A}_{n \ell}^{(\cdot)} = \frac{1}{2\ell+1} \sum_{m=-\ell}^{\ell} (c^{(\cdot)}_{n\ell m})^2\\
  &\mathbf{B}_{n \ell_1 \ell_2 \ell_3}^{(\cdot)} = \sum_{m_1=-\ell_1}^{\ell_1} \sum_{m_2=-\ell_2}^{\ell_2} \sum_{m_3=-\ell_3}^{\ell_3} C^{\ell_3 m_3}_{\ell_1 m_1 \ell_2 m_2} \tilde{c}^{(\cdot)}_{n\ell_1 m_1} \tilde{c}^{(\cdot)}_{n\ell_2 m_2} \tilde{c}^{(\cdot)}_{n\ell_3 m_3} \\
  &\mathbf{T}_{n \ell_1 \ell_2 \ell_3 \ell_4}^{(\cdot)} = \sum_{\ell' \in \mathcal{L}(\ell_1,\ell_2,\ell_3,\ell_4)} \sum_{m'=-\ell'}^{\ell'} P^{(\cdot)}_{n;\ell_1\ell_2;\ell' m'} P^{(\cdot)}_{n;\ell_3\ell_4;\ell' m'} \\
  \text{where}  \quad &P^{(\cdot)}_{n;\ell_a\ell_b;\ell' m'}=\sum_{m_a=-\ell_a}^{\ell_a}\sum_{m_b=-\ell_b}^{\ell_b}C^{\ell' m'}_{\ell_a m_a \ell_b m_b}\tilde{c}^{(\cdot)}_{n\ell_a m_a}\tilde{c}^{(\cdot)}_{n\ell_b m_b}
\end{aligned}
\end{equation*}
where \(C_{\ell_1 m_1 \ell_2 m_2}^{\ell_3 m_3}\) is the Clebsch-Gordan coefficient tensor that couples the three angular momentum components, \(\tilde{c}^{(\cdot)}_{n\ell m} {=} c^{(\cdot)}_{n\ell m} / \sqrt{ \bar{c}_\ell^{(\cdot)}}\) and \(\bar{c}_\ell^{(\cdot)} {=} \frac{1}{n_\text{max} \, (2\ell + 1)} \sum_n^{n_\text{max}} \sum_{m=-\ell}^{\ell} \left(c^{(\cdot)}_{n\ell m}\right)^2 \).

Note that \(\mathbf{A}\), \(\mathbf{B}\), and \(\mathbf{T}\) can be interpreted as the Fourier domain equivalents of the two-, three-, and four-point correlation functions in the spatial domain, \(C_2(\mathbf{r}_1, \mathbf{r}_2)\), \(C_3(\mathbf{r}_1, \mathbf{r}_2, \mathbf{r}_3)\), and \(C_4(\mathbf{r}_1, \mathbf{r}_2, \mathbf{r}_3, \mathbf{r}_4)\).


\end{document}